\documentclass{style/naturep}

\usepackage{authblk}
\usepackage{amssymb, amsmath, graphicx, algorithmicx, algorithm, algpseudocode}
\usepackage{bbm, lineno, ragged2e, setspace, longtable, hyperref}
\usepackage[nameinlink]{cleveref}
\usepackage{anyfontsize, float, booktabs, multirow, xcolor, tabularx, caption, subcaption, geometry}
\usepackage{enumitem}
\usepackage{array}
\usepackage{xspace}
\usepackage{makecell}
\usepackage{threeparttable}

\title{\justifying\bfseries NOAH: Learning the Full Patient Journey. A Longitudinal Multimodal Time-Aware Model for Representation and Forecasting} 

\author[1,2]{Tobias Susetzky}
\author[1, 5]{Raphael Rehms}
\author[1,2]{Dmitrii Seletkov}
\author[1]{Özgün Turgut}
\author[1]{Michelle Espranita Liman}
\author[3,4]{Lisa Steinhelfer}
\author[4]{Rickmer Braren}
\author[1,5,6]{Daniel Rueckert}

\affil[1]{Chair for AI in Healthcare and Medicine, TUM University Hospital and Technical University of Munich (TUM), Munich, Germany}
\affil[2]{Institute for Diagnostic and Interventional Radiology, TUM University Hospital, Munich, Germany}
\affil[3]{Institute for Diagnostic and Interventional Neuroradiology, TUM University Hospital, Munich, Germany}
\affil[4]{Department of Diagnostic and Interventional Radiology and Nuclear Medicine, University Medical Center Hamburg-Eppendorf, Hamburg, Germany}
\affil[5]{Munich Center for Machine Learning (MCML), Munich, Germany}
\affil[6]{Department of Computing, Imperial College, London, United Kingdom}

\newcommand\Heading[1]{\noindent\textbf{\Large{#1}}}
\newcommand\heading[1]{\noindent\textbf{\large{#1}}}
\newcommand{\ours}{\textsc{Noah}\xspace}

\makeatletter
\let\saved@includegraphics\includegraphics
\AtBeginDocument{\let\includegraphics\saved@includegraphics}
\makeatother

\begin{document}
\begin{spacing}{1.2}
\maketitle
\newline
\noindent Corresponding author: Tobias Susetzky (tobias.susetzky@tum.de)

\Heading{Abstract}

\noindent The digitization of healthcare has generated vast, longitudinal, and multimodal patient records over a lifetime, yet fully exploiting these data to represent and predict patient state trajectories remains a critical challenge. Current AI models often struggle to capture the complex, irregular temporal dynamics and inherent stochasticity of real-world multimodal patient data. In the biomedical domain, existing AI approaches for modeling longitudinal patient records are predominantly discriminative, limited to a few modalities, constrained by closed categorical vocabularies, treating time as a monotonic inductive bias, or they are limited in forecasting future patient states. We introduce \textbf{\ours, a time-aware, task-agnostic, generative transformer model representing and forecasting the full multimodal patient journey.} \ours features a novel bidirectional time integration and a variational latent space to capture both the continuous evolution of patient states and the stochasticity of clinical trajectories. Built from over \textbf{559 million clinical events from 431,000 hospital visits of 299,000 patients} across the MIMIC dataset family, \ours natively processes multiple kinds of medical images, time-series and numeric signals, categorical events, as well as structured and unstructured clinical records. \ours is the first truly holistic generative model in its field, enabling autoregressive forecasting with optional time control, zero-shot classification, and counterfactual simulation of clinical interventions. Furthermore, its novel approach generates highly informative and predictive patient state representations that demonstrate strong performance in probing for clinical outcomes, 15 ICD chapters, and 29 comorbidities, as well as in time-to-event prediction. Seamlessly handling diverse modalities and complex temporal dynamics, \ours provides a versatile, task-agnostic, scalable foundation for intelligent predictive systems in personalized clinical care and digital medicine. 

\clearpage
\Heading{Introduction}

Personal health tracking, wearable devices, as well as primary and secondary healthcare systems collect vast amounts of data describing a patient's medical history over time. This comprises vital measurements, medical imaging, textual reports, signals, e.g. from electrocardiograms (ECG), drug prescriptions and intake, medical procedures, diagnoses, and administrative events. The sheer scale, multimodal complexity, and irregular temporal distribution of these lifelong records exceed human personnel capacities by far. Thus, medicine has a critical need for intelligent systems that enable their semantic analysis, at patient and population level, and also predict future disease trajectories. In particular, these systems must accurately reflect the stochasticity of clinical progression where unforeseeable events are omnipresent. By robustly predicting future events and simulating individualized treatment outcomes under specific clinical interventions, such generative models hold the potential to accelerate and improve personalized clinical care and may even exceed human prognostic abilities. For example, such models could support complex scenarios involving multiple subspecialty disciplines (e.g. an acute medical condition superimposed on a chronic disease state), likely to become a more frequent event in an aging society.

With recent AI advances, there is growing interest in exploiting their enormous potential across the medical domain (for instance \cite{medpalm, medgemma}). However, previous approaches largely focus on few isolated modalities (e.g. vision and language \cite{llavamed, medgemma, raddino, biovil-t, medflamingo}, imaging and tabular \cite{lanistr,hager}, or imaging and signals \cite{mmcl,liman,cenikj,selivanov}). Also, they often specialize in specific types of signals \cite{eeg}, require a fixed sampling rate of signals or events, or restrict to inputs at few specific points in time \cite{biovil-t}. In medical representation learning, contrastive approaches have widely proven successful, e.g. \cite{medclip}, albeit subject to the aforementioned restrictions. Simultaneously, transformer-based architectures have been commonly applied e.g. for forecasting \cite{delphi}. However, like natural language models, these approaches usually use inputs consisting solely of categorical data over a closed vocabulary, for instance sequences of ICD diagnosis codes \cite{delphi, steinberg2023motor}. Altogether, even though many promising approaches exist, they fail to holistically capture all of a patient's data and their dynamics in a task-agnostic way that also allows for meaningful, clinical predictions.

Among AI architectures, transformer-based models have proven tremendously successful. Adapting them for irregular clinical event sequences necessitates the integration of temporal position information: current methods commonly apply an additive encoding to a transformer's input tokens or rely on rotary position embedding (RoPE) \cite{rope} to inject temporal information directly into the attention module. RoPE has been adopted in both general-purpose and clinical models, e.g. \cite{tale-ehr}. However, this approach is insensitive to the prediction horizon: a past event may be highly informative for a chronic-disease onset months ahead, yet irrelevant for an acute event in the coming days.

To capture the uncertainty resulting from complex longitudinal dynamics, sequential variational models have historically been the architecture of choice. Variational Recurrent Neural Networks (VRNNs) \cite{vrnn} integrate the principles of Variational Autoencoders (VAEs) \cite{reparametrization-trick-vae}. By providing randomness at inference time, they non-deterministically generate highly structured sequences such as speech or handwriting. In the clinical domain, a patient's state at any given time inherently admits multiple plausible future continuations. Robust forecasting models must explicitly represent this stochasticity. Thus, we adapt this sequential variational concept for a transformer.

Most recent advances show that including a patient's long-term history can improve a model's diagnostic and prognostic abilities (cf. \cite{tamme, delphi}). They have established flexible token representation methods for multimodal patient data: \textsc{Tamme} \cite{tamme} considers health records a sequence of timestamped events with both continuous and discrete information, and represents them jointly in a shared space. This makes them easily accessible for transformer-based classification models and unsupervised pretraining via random masking and reconstruction. Scaling up \textsc{Tamme}'s token representation methodology and pretraining strategy, \textsc{Apollo} \cite{apollo} demonstrates its real-world applicability to representation learning and clinical downstream usage. At inference time, it uses a learned token tied to a specific event type, e.g. diagnosis, to prompt the model to produce one patient representation vector. However, approaches in this family are, at their core, discriminative, lacking the ability to fully reflect temporal dynamics and sequential inter-event dependencies, as well as the capacity for true forecasting and generative application with a source of variability. Neither their architecture nor their training objective directly encourages learning clinical dynamics and progression.

To overcome these limitations, we introduce \ours, a \textbf{time-aware, task-agnostic, generative, multimodal} transformer model designed to learn a \textbf{latent space of patient states and transitions} from lifetime sequences of multimodal patient data. We adapt a \textbf{context-conditional variational approach} to explicitly capture the stochasticity of disease progression, enabling the model to learn predictive patient state representations over time while accounting for unexpected transitions. Notably, \ours fully consumes the actual \textbf{content of all modalities}, e.g. imaging data or waveforms, while providing \textbf{variation at inference time} when autoregressively forecasting the next events. With a novel \textbf{bidirectional time integration} \ours can natively handle irregular inter-event time gaps and weight the relevance of elements in a patient's history for the next prediction based on both recency and prediction horizon.

We demonstrate that \ours serves as a highly versatile task-agnostic foundation for holistic patient trajectory analysis. Pretrained and evaluated on \textbf{559 million timestamped clinical events} across the MIMIC dataset family \cite{mimiciv, mimic-ecg, mimic-echo, mimic-note, mimiccxr, mimiced}, \ours successfully captures the temporal dynamics of patient journeys. We validate its performance across a diverse suite of clinical downstream tasks, demonstrating its efficacy in probabilistic \textbf{zero-shot forecasting, time-to-event prediction}, and generating compact \textbf{patient state representations}. Finally, we highlight its generative capabilities for autoregressive \textbf{rollouts with optional time control} and \textbf{counterfactual} simulations of clinical interventions, providing a scalable architecture for hypothesis generation and in silico trial design.

Overall, with \ours, we introduce a ready-to-scale architecture for a foundational holistic patient trajectory model and demonstrate its potential and versatile applicability.


\clearpage
\Heading{Results} 
\vspace{1em} 

\begin{figure}[ht!]
    \centering
    \includegraphics[width=\linewidth]{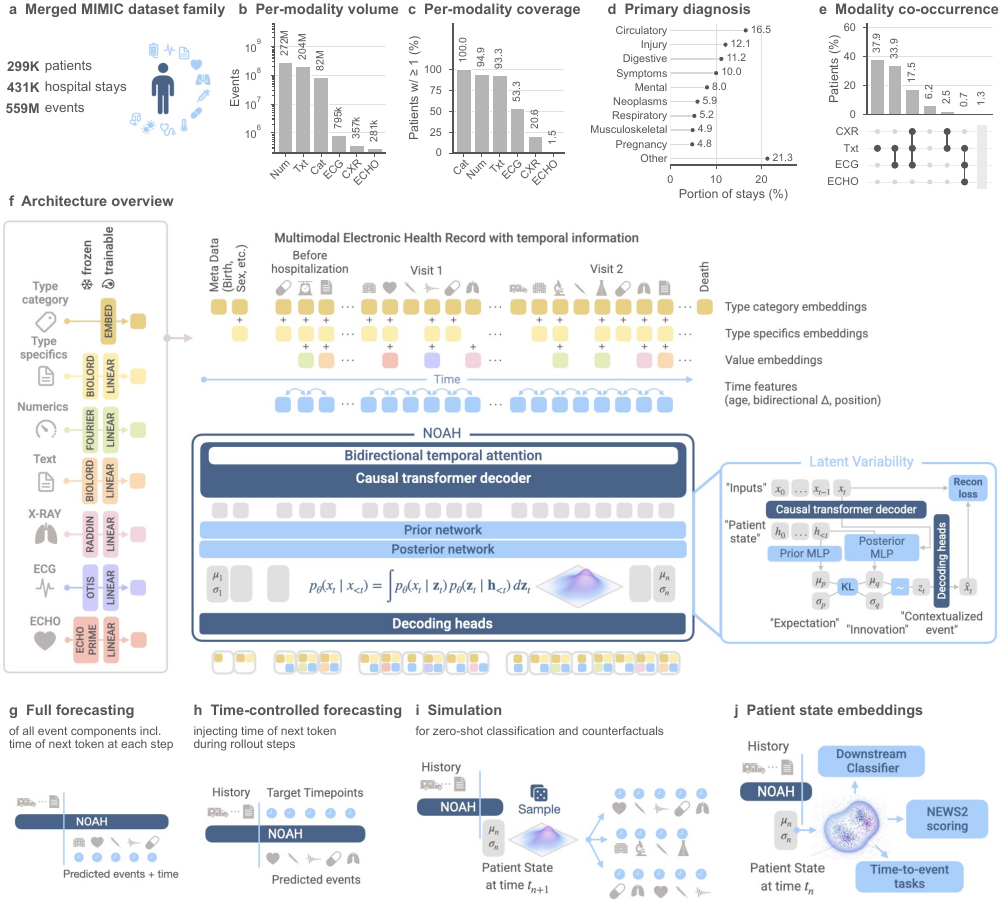}
    \caption{\footnotesize \textbf{\ours understands, represents, and extends multimodal patient records.} (a-e) Patient journey dataset built from MIMIC. (f) \ours architecture: multimodal health record events embedded through type category, type specifics, and value. A transformer decoder pretrains on these sequences with causal masking and novel time integration. A novel latent module explicitly learns context-conditional patient states and transitions. (g-h) Autoregressive rollout with optional time control. (i) Outcome simulation for Monte Carlo-style classification and counterfactual analysis. (j) Downstream applications of \ours's patient state embeddings.}
    \label{fig:1}
\end{figure}

\section*{A time-aware model can understand and predict multimodal data over the entire patient journey.}
    \ours is a transformer model explicitly trained to learn temporal dynamics of patient states across multimodal trajectories (Fig.~\ref{fig:1}). It introduces a bidirectional temporally enriched attention mechanism to deeply incorporate time features and offer temporal control during generative inference. \ours is pretrained and evaluated on 559M timestamped events from 299k patients spanning 431k hospital visits (panels \textbf{a}-\textbf{e}) as a versatile foundation for numerous downstream setups in a real-world clinical context. It operates on multimodal event sequences comprising texts, X-ray and echocardiogram (ECHO) imaging, electrocardiogram (ECG) waveforms, numerical and categorical data, in arbitrary order and count (\textbf{f}). Event representations combine high-level type category, free-text type specifics, externally obtained value embedding, and temporal features. Free-text specifics and continuous values overcome limitations of a closed event vocabulary and enable a trained model to consume unseen data such as updated drug names. By integrating timing information for each event such as current age and bidirectional inter-event deltas, \ours naturally works on temporally irregular sequences. As a key novelty, \textbf{\ours is fully multimodal, yet generative}. It is a transformer decoder offering variability during autoregressive inference while maintaining a coherent patient state across output components of each timestep. The pretrained model can be directly used for representation, i.e. to obtain lifetime trajectories as sequences of patient state embeddings, and for extending them to forecast future developments (\textbf{g}-\textbf{j}).

\begin{figure}[ht!]
    \centering
    \includegraphics[width=\linewidth]{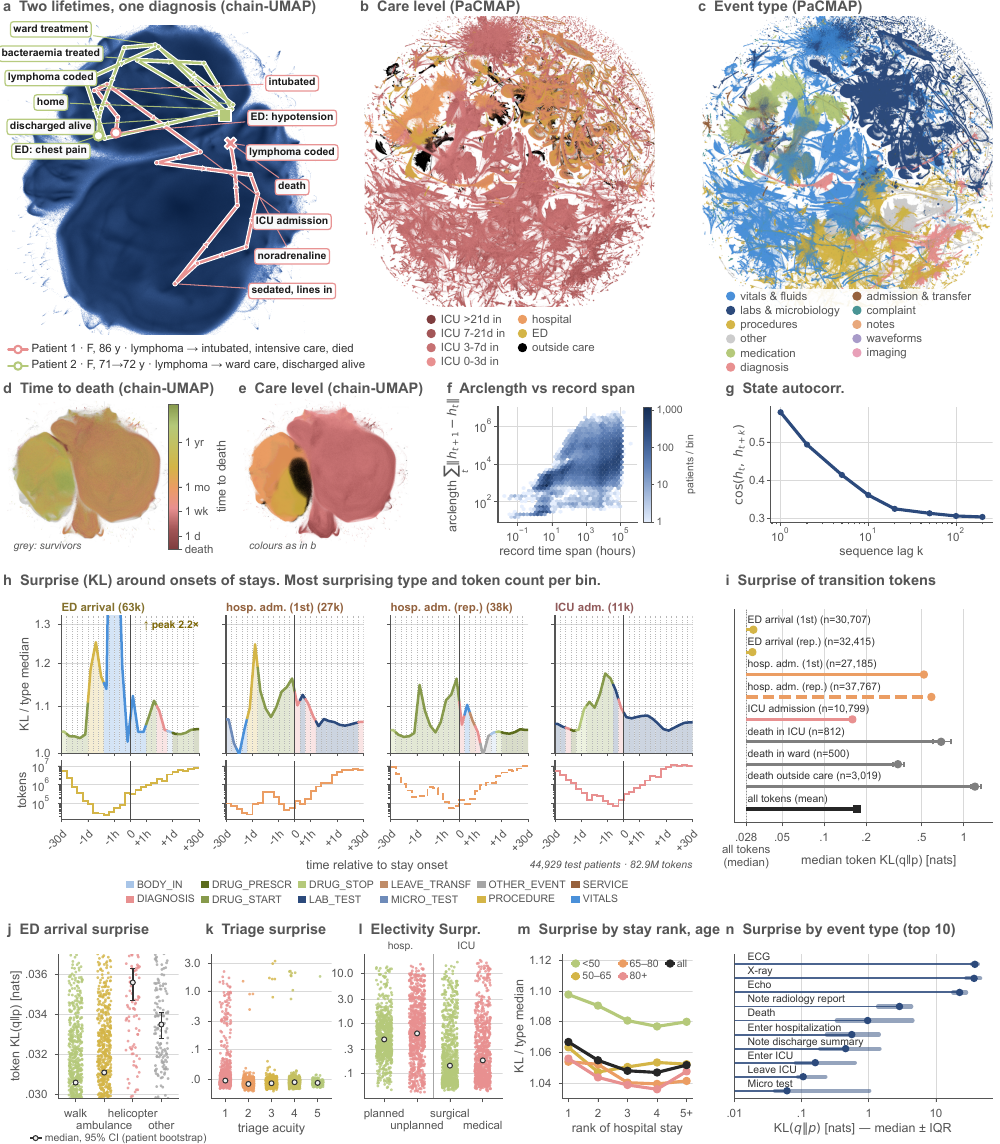}
    \caption{\footnotesize \textbf{\ours learns patient trajectories as states and transitions.} (a-c) All patient states in latent space, mainly organized by contextualized event types. (a) Two exemplar trajectories with similar starting conditions, yet diverging course in embedding space. (d-e) Regions of care levels and risk (time until death). (f-g) Trajectories' exploration of latent space. (h-n) The model's surprise at state transitions near stay onsets and across different event types correlates with clinical novelty.}
    \label{fig:2}
\end{figure}

\section*{\ours learns patient trajectories, a landscape of states and transitions.} 
    We feed \ours the entire record of each test patient, extract the 768-dimensional transformer output before the variability module as \emph{patient state representation} for each timestep, and investigate the resulting trajectories (Fig.~\ref{fig:2}): they unfold and evolve over time in the embedding space (\textbf{f}-\textbf{g}). Qualitatively, patients with similar starting conditions, yet different course and outcome, diverge in this space (\textbf{a}). Unlike static atlases of isolated event types, the learned space is primarily organized by \emph{contextualized events}, e.g. by event type under care level and clinical progress (\textbf{b}, \textbf{c}). Albeit subject to projection effects, panels \textbf{d}-\textbf{e} indicate a clear separation by care intensity, e.g. long-term ICU vs. general hospital, emergency department (ED), and outside care (\textbf{e}). Lower-intensity care corresponds to the stable, non-critical risk region (\textbf{d}), while a high-risk zone emerges close to the ICU region. For these 2D projections, we use PaCMAP \cite{pacmap} and chain-UMAP, a UMAP customization \cite{umap} ensuring a token's sequence neighbors are in the UMAP nearest neighbors. This preserves within-patient structure without affecting inter-patient and global insights. 
    
    As a characteristic of \ours, we can quantify the model's surprise in patient state transitions, i.e. the difference between prior ("expectation") and posterior for the next token. We find that this surprise often corresponds to clinical novelty: surprise increases shortly before the formally charted stay beginnings (\textbf{h}). It peaks for ED vitals, followed by drug administration and immediate procedures in other care. This reflects real unpredictability of acute events and first responses, especially in first-time visits. The model reacts to abnormal outpatient measurements that precipitate the stay, before formal transition. Surprise decreases with more available context, stabilizing patient condition, and emerging routines. Among transition events themselves, ED arrival is well predictable due to charting patterns, while death outside clinical settings due to acute events is hardly predictable (\textbf{i}). Overall, \ours's surprise coincides with clinical reality. This also holds for ED arrival and acuity (helicopter vs. walk-in, acuity 1 vs. rest), electivity of procedures (planned vs. unplanned), and increasing predictability with more preceding visits and higher age (\textbf{j}-\textbf{m}). Genuinely hard to predict are content-rich modalities such as ECG and X-ray (\textbf{n}, Supplementary Table~\ref{tab:latent_kl_by_event_type}).

\begin{figure}[ht!]
    \centering
    \includegraphics[width=\linewidth]{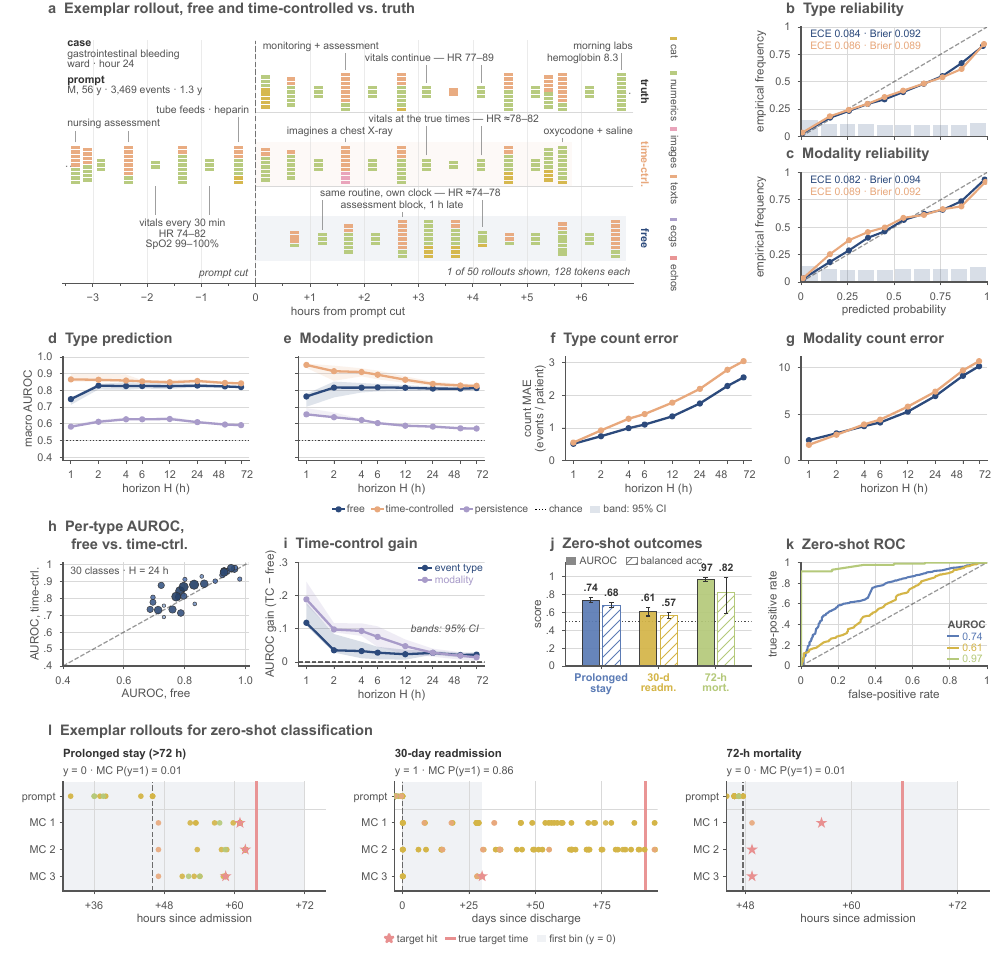}
    \caption{\footnotesize \textbf{\ours can autoregressively roll out future trajectories to make clinical forecasts.} (a) Autoregressive next token prediction for an exemplar patient, with the actual time delta to the next token provided (\emph{time-ctrl.}) and without (\emph{free}). (b-g) Model calibration and prediction quality for type categories and modalities. (h-i) Effect of time control on performance in terms of occurrence of each type category and modality within a certain temporal horizon. Panel (h) shows results for event type only. (j-l) Zero-shot classification results and examples of \ours simulating multiple possible trajectory continuations per patient and then using Monte Carlo estimation for probabilities of clinical outcomes.} 
    \label{fig:3}
\end{figure}

\section*{\ours can forecast patient trajectories with optional time control.} 
    For $13{,}551$ eligible test patients, we prompt \ours with their lifetime sequences up to a certain point (Fig.~\ref{fig:3}\textbf{a}). The frozen model then autoregressively rolls out future events to fill the held-out window up to different temporal horizons. In a Monte Carlo fashion, we estimate the probabilities and counts of each event type category and modality to occur within each horizon. We compare against the held-out ground truth (Extended Data Table~\ref{tab:forecasting-headline}, Supplementary Tables~\ref{tab:forecasting-los-subgroup}, \ref{tab:forecasting-perclass}). \ours achieves AUROC scores of $0.83$-$0.95$ with Brier scores $0.03$-$0.11$, outperforming a naive persistence baseline that repeats the pre-anchor time window (\textbf{b}-\textbf{e}). The model is reliable with ECE $0.08$ to $0.094$, yet with slight overconfidence at high predicted probabilities (\textbf{b}-\textbf{c}). It yields MAEs of $\leq 3.1$ and $\leq 10.7$ occurrences on average for type and modality classes, naturally increasing within 72 h, as errors propagate during rollout and divergence from ground truth increases (\textbf{f}-\textbf{g}). Within these experiments, providing the ground truth time delta to the next token (\emph{time-control}), i.e. enforcing the time at which \ours should predict the next event in each step, improves performance significantly, up to $+0.19$ AUROC within shorter, more sensitive horizons (\textbf{h}-\textbf{i}). However, unlike the error in the predicted number of modality occurrences, the error in type count actually increases for longer horizons under time control (\textbf{f}-\textbf{g}). We attribute this to errors in other sensitive event components (e.g. the value embedding) where enforced timing moves less accurate predictions even further out of the learned distribution.

\section*{Monte Carlo simulation enables zero-shot clinical decisions.}
    \ours's autoregressive rollout also enables Monte Carlo estimation of clinical outcome probabilities: we perform binary zero-shot classification for prolonged stay, mortality, and readmission. Prompting again with the full patient history including the beginning of the current stay, we draw $M\in\{25, 50\}$ trajectories, locate the first occurrence of the target token in each (e.g. a discharge event for the length-of-stay task) and bin its time delta to the reference event, which is either admission or discharge depending on the task (Fig.~\ref{fig:3}\textbf{l}, Extended Data Fig.~\ref{fig:res_zs_ex}). Under a prompt cutoff at 48h after admission, we achieve an AUROC of $0.97$ (Balanced Acc. $0.82$) for 72h mortality and AUROC $0.74$ (Balanced Acc. $0.68$) for length-of-stay $\geq 72$ h prediction. Prompting the model with events up to discharge (exclusive), \ours only slightly separates positives for 30-day readmission with an AUROC of $0.61$ (Balanced Acc. $0.57$). We attribute this to error propagation and distribution drift over long-term rollouts, since training only predicts one next token instead of multiple autoregressively. The difficulty of this task also matches clinical intuition. For Monte Carlo estimation we discard rollouts without a scorable outcome, resulting in a coverage of $65$\% for this long-term target, yet $80$-$82$\% for the significantly shorter simulations for prolonged stay and mortality. Detailed results in Fig.~\ref{fig:3}\textbf{j}-\textbf{k}, Extended Data Table~\ref{tab:zeroshot}.
    
\begin{figure}[ht!]
    \centering
    \includegraphics[width=\linewidth]{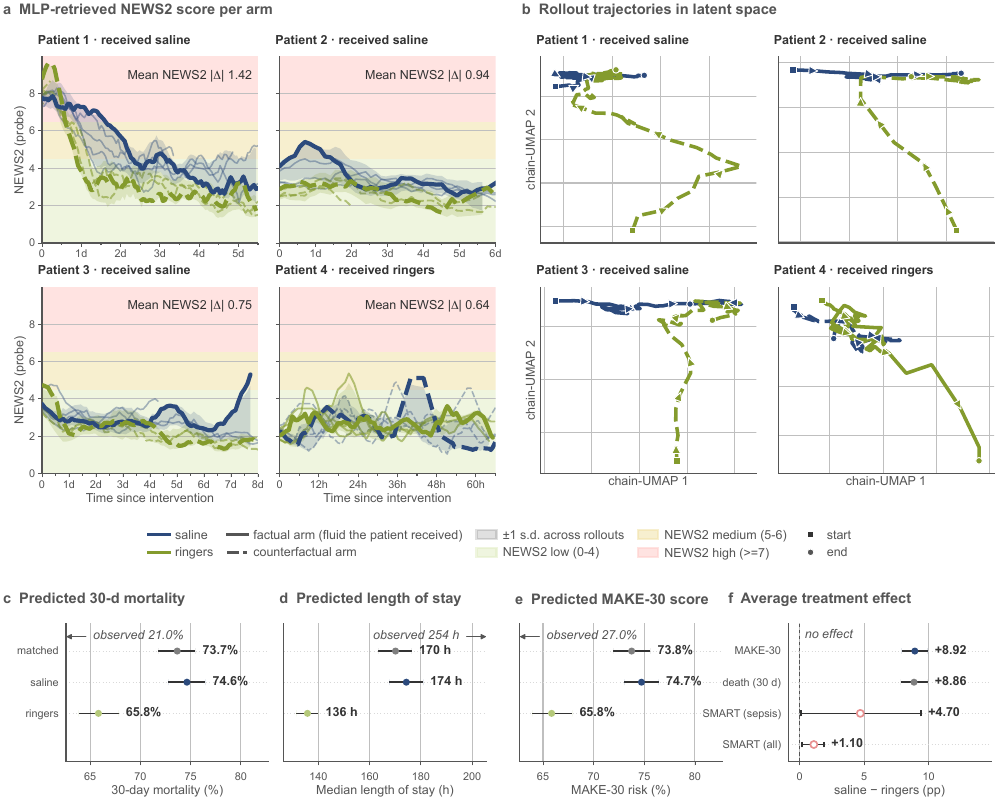}
    \caption{\footnotesize \textbf{\ours simulates trajectories and effects of counterfactual interventions.} For sepsis patients, we prompt \ours with the patient's history followed by the administration of either saline or Ringer's and then estimate outcomes via Monte Carlo simulation before comparing the simulated treatment effect to the sepsis subgroup of the SMART clinical trial \cite{counterfactual-medical-study-sepsis}. (a) NEWS2 score retrieved from generated patient states during autoregressive rollouts. (b) Projections of those patient state trajectories in latent space via chain-UMAP projection as used and described for Fig.~\ref{fig:2}. (c-e) Lower predicted mortality, shorter length of stay, and lower MAKE-30 score under the administration of Ringer's compared to saline. Yet, the model overpredicts mortality and underpredicts stay duration compared to the observed factual outcome. (f) Simulations show an average treatment effect in clear favor of Ringer's, matching the sign of the effect in SMART's sepsis subgroup at roughly twice its magnitude.}
    \label{fig:4}
\end{figure}

\section*{\ours enables counterfactual simulations of clinical intervention.} 
    \ours can also simulate the effect of counterfactual interventions. We demonstrate this in one high-relevance case that allows direct comparison with a clinical trial: \ours estimates in-hospital mortality, length of stay, and MAKE-30 (see \cite{counterfactual-medical-study}) for sepsis patients under two arms, the factual and counterfactual choice of 0.9\% saline vs. lactated Ringer's as the first intravenous fluid within six hours after a sepsis marker. We prompt \ours with the patient's events truncated after this intervention, for one arm swap the factual fluid for the counterfactual, then for both arms perform autoregressive rollouts and Monte Carlo estimation of outcomes as described above. We observe simulated treatment effects of $+8.92$ and $+8.86$ points for MAKE-30 and death for saline vs. Ringer's (Fig.~\ref{fig:4}\textbf{c}-\textbf{f}), matching the finding and effect sign in SMART's sepsis subgroup \cite{counterfactual-medical-study-sepsis} at roughly twice its magnitude (Extended Data Table~\ref{tab:counterfactual-headline}, Supplementary Tables~\ref{tab:counterfactual-calibration}, \ref{tab:counterfactual-coverage}, \ref{tab:cf-make30}). \ours overpredicts mortality (and thus MAKE-30) in this setting. This effect grows with increasing prediction horizon. We attribute it to autoregressive rollout drift leaving the learned distribution: a clean in-distribution trajectory of a survivor requires a long-term rollout not emitting significant clinical events. This is unlikely by design. However, we stress that this overprediction does not affect \emph{within}-patient contrast. We further investigate trajectories qualitatively and observe that arms diverge, consistent with the simulated treatment effect (\textbf{a}-\textbf{b}).

\begin{figure}[ht!]
    \centering
    \includegraphics[width=\linewidth]{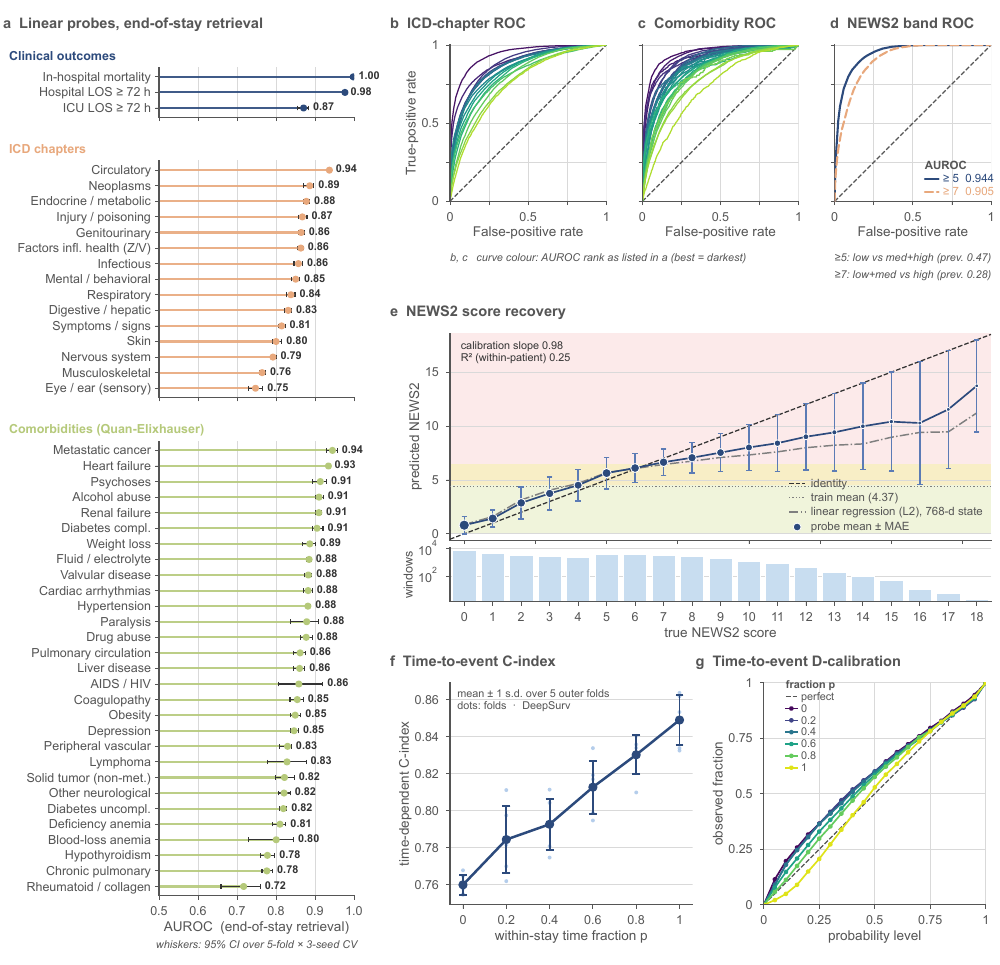}
    \caption{\footnotesize \textbf{\ours produces task-agnostic patient state embeddings for multiple downstream applications.} (a-c) Linear probing for clinical information retrieval across 3 outcomes, 15 ICD chapters, and 29 comorbidities at the end of a stay. (d-e) Probing for the NEWS2 patient stability score (d) at end of stay and (e) at each complete local window of the vitals required for NEWS2. Bar chart indicates counts of windows. (f-g) The td-C-index and D-calibration for time-to-event survival models trained across the duration of a patient's stay.} 
    \label{fig:5}
\end{figure}

\section*{\ours's patient state representations carry clinical information and risk over time.}
    We probe \ours's patient state embeddings for clinical information, fitting an $\ell_2$-regularized logistic regression under stratified five-fold cross-validation over three seeds. For retrieval from a stay's final patient state, we achieve AUROC scores within $[0.72, 0.94]$ for 15 ICD chapters and 29 comorbidities (Quan-Elixhauser \cite{quan2005coding}), and $0.87$, $0.98$, and $>0.99$ for ICU length of stay, hospital length of stay, and mortality (Fig.~\ref{fig:5}\textbf{a}-\textbf{c}). When probing at different points over time, we observe peak performance at the beginning of the stay for most tasks (Extended Data Fig.~\ref{fig:res_lp_tmp}, Supplementary Tables~\ref{tab:lp-outcomes}, \ref{tab:lp-icd}, \ref{tab:lp-elix}). At this point, signals such as complaints and initial treatment are strongest, while in mid- and long-term care, routine procedures dominate (Extended Data Fig.~\ref{fig:ds_instay}). We observe better results on younger patients and shorter stays (Supplementary Fig.~\ref{fig:res_lp_cohort}, Supplementary Table~\ref{tab:lp-cohort}) and conclude that the embeddings "forget" over time, focusing on predictive features more than summarization. This is deliberate given the model's nature. In addition to these probes, we train a regression MLP to retrieve the NEWS2 score \cite{news2} for patient state severity on a scale from 0 to 20 points, covering low (0-4), medium (5-6), and high (7+) risk. We compute ground truth deterministically on patient timelines after aggregating the necessary vitals within 1h windows. On $39{,}952$ windows from $4{,}414$ held-out patients, we achieve an MAE of $1.34$ and AUROC values of $0.94$ and $0.91$ for separating low vs. medium-high and low-medium vs. high (\textbf{d}, Supplementary Tables~\ref{tab:news2-headline}, \ref{tab:news2-breakdown}). The NEWS2 regression is well calibrated on the low and medium levels, only diverging from the ground truth on rarer high-risk scores (\textbf{e}, Extended Data Fig.~\ref{fig:res_lp_risk}).

\section*{\ours's patient state representations enable survival analysis.}
    We assess the survival analysis capabilities of \ours's patient state representations from the last stay, denoted $\mathbf{h}_{<pT}$, as more within-stay longitudinal context becomes available. We evaluate this by predicting survival from embeddings indexed at increasing within-stay time ($T$) fractions ($p \in \{0.0, 0.2, 0.4, 0.6, 0.8, 1.0\}$, where $p=0.0$ is entering the hospital, and $p=1.0$ is the last event before death or discharge) using the same patients and endpoint (time-to-death) across all settings. The results demonstrate that \ours's embeddings can be effectively used for survival analysis (Fig.~\ref{fig:5}\textbf{f}-\textbf{g}). Performance improves as embeddings from later phases of the patient's stay are incorporated, leading to higher discrimination in terms of the time-dependent C-index~\cite{antolini05}, while maintaining well-calibrated predictions in terms of D-calibration~\cite{haider20, sabook}. These findings suggest that progressively richer longitudinal patient representations provide increasingly informative signals for time-to-event prediction.

\begin{figure}[ht!]
    \centering
    \includegraphics[width=\linewidth]{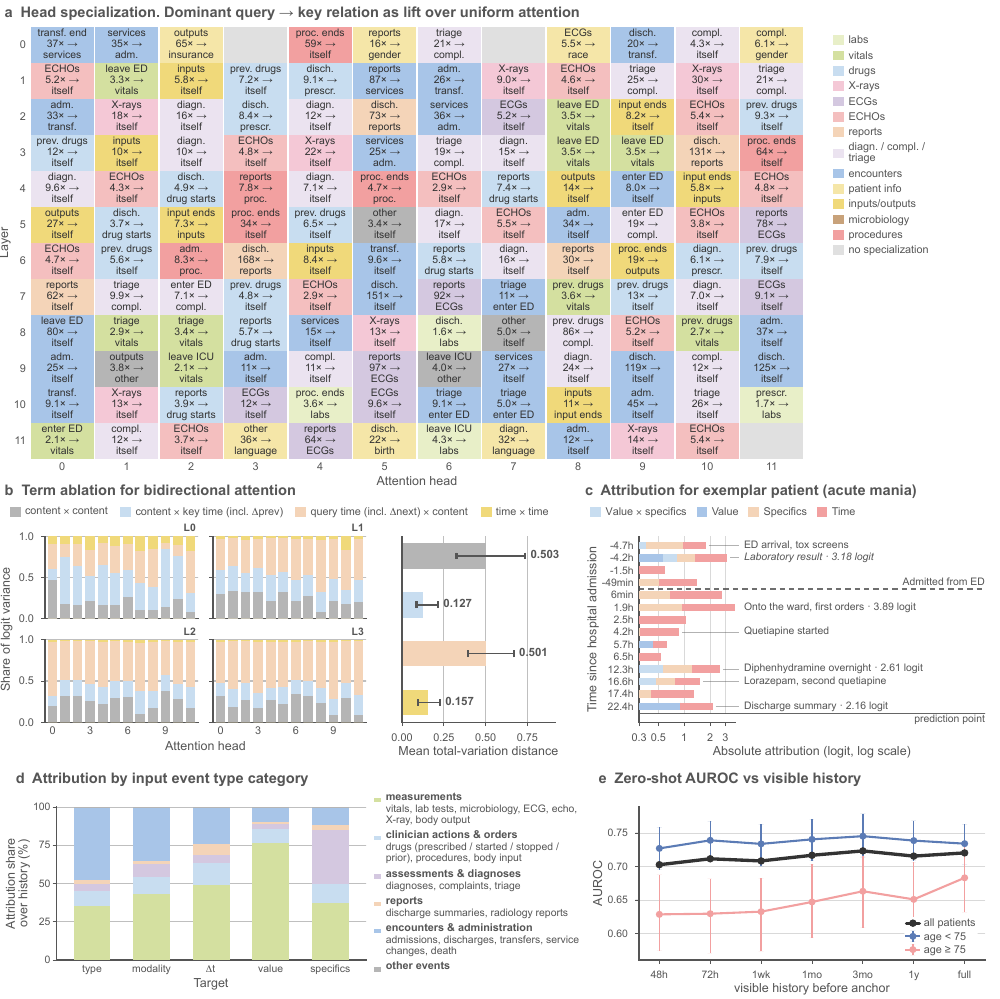}
    \caption{\footnotesize \textbf{\ours is transparent: attention distribution and input attribution.} (a) The model's attention heads specialize in clinically relevant relations, capturing within-type longitudinal dynamics and dependencies between observations and charting. The numeric value in each cell is the lift over a uniformly distributed attention. (b) Ablation of single terms in bidirectional temporal attention. Time features on the queries including the forward time delta matter most after the first layer in which the model can merge historic time dynamics into the content channels. (c) Exemplar attribution over a patient timeline for predicting the next token. Obtained using Integrated Gradients with "event type"-only as baseline. For this case of acute mania, the model attends highly to lab results, admission, and medication. (d) Attribution of event types by target. Patient measurements matter, especially for predicting the next value component. Administrative info is most important for event type and never outweighs other information. This illustrates the model actually learns from physiology, not just structure. (e) Ablation of patient history for zero-shot prediction of prolonged hospital stay. History inclusion matters significantly, yet mostly for elderly patients.}
    \label{fig:6}
\end{figure}

\section*{\ours relies on time and specializes in clinical channels.}
    We investigate the internals of the pretrained \ours model: Of 144 attention heads, 141 specialize in clinical relations (Fig.~\ref{fig:6}\textbf{a}): we measure each query $\rightarrow$ key relation's attention share lift over uniform and identify the dominant one. 76 heads relate events of one type to past events of the same type, e.g. X-ray $\rightarrow$ X-ray for earlier studies. The model maintains per-modality longitudinal context. 65 heads connect observation and reporting, e.g. notes $\rightarrow$ ECG, leaving ED $\rightarrow$ vitals. The model has internalized workflows and documentation logic. For explainability, the model allows analyzing which elements in a patient's record the prediction rests on (\textbf{c}, Extended Data Fig.~\ref{fig:res_ip_ex}). Attribution decomposition by event types shows \ours relies on physiology, not just structure (\textbf{d}): patient measurements are crucial for predicting the next token, while clinical structure is most important for the type component yet never outweighs the other inputs. Further, we investigate importance of temporal context: our bidirectional temporal attention decomposes into four terms: content-content (CC), time-content (TC, query including time delta to next event), content-time (CT, key including the time delta to previous event), and time-time (TT) attention. We ablate each and measure total variation distance between resulting attention distributions: attention shifts most under omission of CC and TC (\textbf{b}). For CC this is trivially expected. TC being dominant shows the model attends to history differently depending on the prediction horizon. CT is dominant only in the input layer, before the model merges content and time channels (\textbf{b}). The forward time delta can never arise from history under causal masking, so it must be carried explicitly. We also ablate patient history in our zero-shot prolonged hospital stay setting with $M=25$ rollouts: with more history AUROC rises from $0.63$ to $0.68$ for elderly patients. Overall, history improves AUROC from $0.70$ to $0.72$, while the model performs better on younger subjects for all extents of history (\textbf{e}).
 \newpage
\clearpage
\Heading{Discussion}

With \ours, we developed a multimodal, time-aware, generative model that learns longitudinal dynamics of lifetime patient journeys for representation and forecasting. We introduced a novel bidirectional time integration and a variational latent space of context-conditional patient states and transitions. The latter provides variability at inference time while maintaining semantic coherence across components of the generated events. We demonstrated \ours's ability to learn and represent patient states over time and verified the usability of these embeddings: lightweight linear models retrieve clinical information and stability scores from highly informative features at any point in time, eliminating the need to directly process complex patient data in downstream systems. Time-to-event models prove these task-independent features highly predictive, underlining their prognostic value. We also established the model's autoregressive capability for forecasting future clinical events with optional time control. This enables grasping likely developments in the near future easily and may serve as a flexible prediction engine. Here, we estimated probabilities of clinical outcomes in a zero-shot fashion, and simulated treatment effects under counterfactual interventions.

\ours is the first model to holistically capture patient states and their transitions, incorporating any modality acquired in the clinical setting, entire history, and temporal dynamics. Unlike previous approaches, \ours provides a continuous holistic patient state representation at any point in time, forming semantically meaningful trajectories that the model itself can even extend in a task-agnostic way to simulate any future developments.

However, \ours comes with the exposure bias of autoregressive models: it trains predicting one token ahead under teacher forcing with ground truth predecessors, while at rollout it iteratively consumes re-encoded predictions. This propagates errors for long horizons, yet might be overcome by sufficiently large reliable training data and post-training techniques. Also, we stress that \ours's representations should not be considered \emph{summaries} of the patient history. The model's objective is to extract features that are \emph{predictive}. Further, we demonstrate the successful application of \ours to a \emph{broad} range of tasks. For feasibility, we focus on selected use cases within them, while each task in general constitutes a challenging problem on its own. Future work may deepen these experiments beyond our scope. Finally, \ours is not a physiologically grounded deterministic patient simulator, but a probabilistic model capturing statistical patterns. It depends on training data and estimates likely developments, rather than deducing logically. Counterfactual simulations are not causal modeling as known from other fields, but intervention-conditioned stochastic forecasts.

In summary, \ours's novel approach provides a medical foundational model architecture to serve as both \textbf{one universal patient state encoder and a prediction engine across all modalities and events in a patient's lifetime}. Its holistic patient view and generative approach surface risks and future developments to clinicians and patients alike. It opens paths to rethink the utilization of large-scale lifetime records currently collected worldwide. A \ours-based foundation model built from these will impact AI integration in the medical domain and flexibly facilitate clinical care, including simulation, risk prediction, and interaction with patient history. It has the potential to significantly improve personalized treatment and digital medicine, while making the rich information of lifelong multimodal health data accessible to a broad audience.

\clearpage
\Heading{Online Methods}
\vspace{1em}
\newline
\heading{Dataset and preprocessing}

We merge all of the following subsets from the Medical Information Mart for Intensive Care (MIMIC) from PhysioNet \cite{physionet} on subject IDs to obtain a large-scale dataset of multimodal patient timelines: MIMIC-IV 2.2 \cite{mimiciv}, MIMIC-ED 2.2 \cite{mimiced}, MIMIC-Note 2.2 \cite{mimic-note}, MIMIC-CXR-JPG 2.0 \cite{mimiccxr}, MIMIC-IV-ECG 1.0 \cite{mimic-ecg}, and MIMIC-ECHO 1.0 \cite{mimic-echo}. Every charted data point of these multimodal health records, i.e. every single piece of information about a patient, is treated and processed as a standalone timestamped event. For ongoing processes such as a surgical procedure covering a certain time span, dedicated end tokens are inserted on the patient timeline at the respective timestamp. In this way, they implicitly provide the duration information to the model without the risk of leaking it at an earlier timestep. Charting frequencies of event types can be highly disproportionate, especially during ICU stays where vital measurements are collected automatically at short, fixed intervals with minimal informational gain. To limit their structural dominance over other events, a sliding window aggregates numeric vital measurements and represents each segment by its minimum and maximum for each type of vitals, timestamped at the window mean. We split our overall dataset into training, validation, and test sets according to a 70/15/15 ratio under the constraint that every patient appears in exactly one split, preventing data leakage. In our resulting dataset, one sequence corresponds to the lifetime records of one patient, starting at birth. It does not necessarily end with death and will usually include large temporal gaps as well as arbitrarily many ED visits, hospitalizations, and ICU stays, but also data from online medical records, and e.g. medication taken outside clinical care. For training and validation, each lifetime sequence is segmented into sliding windows of at most $2048$ tokens with an overlap of $256$. Every window is prepended with a fixed demographic prefix (date of birth, race, gender, insurance, language, and marital status). The latter four reflect the most recent value available at the window end and fall back to ``Unknown'' where missing. A mask excludes this metadata prefix and the window overlap region from the training objective. Each event after the prefix contributes to the loss exactly once. Patients with fewer than 64 of these scored tokens are excluded to ensure a minimum of informative history to learn from. In the test set, each patient is represented by a single uncapped sequence spanning their full record, with every token evaluated. Unlike existing work, sequences exceeding the model's input length are not invalidated by subsampling, but instead processed in consecutive chunks at inference time with the meta prefix prepended for each chunk. For performance reasons, we precompute the value embeddings, i.e. the representations of images, texts, electrocardiography (ECG), and echocardiography (ECHO), as well as type-specifics embeddings obtained from external pretrained models. For images we use RAD-DINO \cite{raddino}, while for textual data we rely on BioLORD \cite{biolord}. We use ECG data from the MIMIC-IV-ECG corpus containing $800{,}035$ $10$-second, $12$-lead recordings sampled at $500\,$Hz. We extract one ECG embedding per recording using the frozen \textsc{OTIS}~\cite{otis} model, a tiny general-purpose time series encoder pretrained on a multi-domain time series corpus. To this end, each ECG recording is partitioned into non-overlapping $1 \times 24$ patches by a strided convolution, yielding a sequence of $12 \times 208 = 2496$ tokens of dimension $192$. These input tokens are processed by \textsc{OTIS} and the output patch tokens at the final layer are mean-pooled and used as the $192$-dimensional global embedding of the ECG recording. From MIMIC-ECHO 1.0, we include $281{,}059$ echocardiogram videos. Embeddings for these videos are generated using the EchoPrime model \cite{echoprime}. The global \texttt{[CLS]} token is used as the embedding representation for each echocardiogram. Prior to embedding, each echocardiogram is preprocessed by first extracting the ultrasound region using the DICOM metadata SequenceOfUltrasoundRegions, followed by removing the ECG overlay and any remaining burned-in annotations (e.g. text). Videos are then resized to $224 \times 224$ and saved as AVI files. Following EchoPrime, during data loading, a further 10\% zoom crop is applied to reduce empty background before resizing back to $224 \times 224$. Pixels are normalized using the mean and standard deviation calculated from the EchoPrime train set. Additionally, each echocardiogram video is sampled to 16 frames at a stride of 2, with shorter videos zero-padded to a uniform length. For cohort and dataset statistics, see Extended Data Table~\ref{tab:dataset-cohort-demographics}, Supplementary Table~\ref{tab:dataset-modality-statistics}, and Extended Data Fig.~\ref{fig:ds_instay}.


\heading{Model and pretraining}
\paragraph{Overall architecture and pretraining.} 
    We pretrain a causally masked transformer to learn sequences of patient state representations $(\mathbf{h}_t)_{t\in\{1,...,S\}}, \mathbf{h}_t \in \mathbb{R}^d$. For this, every sample in our dataset is a temporally ordered sequence $(\mathbf{x}_t)_{t\in\{1,...,S\}}$ of highly heterogeneous events, each with type and time information, as well as an optional continuous value embedding obtained from an external model (see Section~\textbf{Dataset and preprocessing} above). We employ trainable projection modules and a composite representation scheme to transform each sample into a sequence of uniform tokens $(\mathbf{e}_t)_{t\in\{1,...,S\}}, \mathbf{e}_t\in\mathbb{R}^d$ in a shared space, see Section~\textbf{Multimodal token representation} below. To integrate the temporal information deeply into the model, we encode it at two points: directly at input tokens like a conventional position encoding, and in the self-attention modules, where backward-oriented features are added to the keys and forward-oriented features to the queries (cf. Section~\textbf{Bidirectional temporal attention}). To reconstruct event components (type, time, value embedding, etc.) from the transformer outputs, we add a dedicated decoding head per component (Section~\textbf{Multimodal output decoding}). For variability at inference time without negatively impacting semantic coherence across these output components (i.e. a valid patient state where type, value, etc. in combination are realistic and plausible), we add a latent patient state space between transformer outputs $(\mathbf{h}_t)_{t\in\{1,...,S\}}$ and decoding heads: we exploit causal masking and construct a contextual prior $p_\theta(\mathbf{z}_t \mid \mathbf{h}_{t-1})$ from the previous timestep's transformer output $\mathbf{h}_{t-1}$ with parameters $\theta$ and an informed posterior $q_\phi(\mathbf{z}_t \mid \mathbf{h}_{t-1}, \mathbf{e}_t)$ with access to the current timestep's input $\mathbf{x}_t$ via its embedding $\mathbf{e}_t$, using parameters $\phi$. For $t=1$, $\mathbf{h}_0$ is given by the metadata prefix of the sequence. The model's objective during self-supervised pretraining is to reconstruct $\mathbf{x}_t$ from the posterior-sampled $\mathbf{z}_t$ while simultaneously minimizing the divergence between posterior and prior, i.e. the Kullback-Leibler (KL) divergence:
    \vspace{0.05cm}\\
    \[
    D_{\mathrm{KL}}\bigl(q_\phi(\mathbf{z}_t \mid \mathbf{h}_{t-1}, \mathbf{e}_t) \,\big\|\, p_\theta(\mathbf{z}_t \mid \mathbf{h}_{t-1})\bigr) = \mathbb{E}_{\mathbf{z}_t \sim q_\phi}\!\left[\log \frac{q_\phi(\mathbf{z}_t \mid \mathbf{h}_{t-1}, \mathbf{e}_t)}{p_\theta(\mathbf{z}_t \mid \mathbf{h}_{t-1})}\right]
    \]
    \vspace{0.05cm}\\
    This directly encourages the model to learn an informative history representation, limited by the surprise of unexpected events, the \emph{innovation} (see Section~\textbf{Variational autoregression}). 

\paragraph{Multimodal token representation.}
    Like \textsc{Apollo} \cite{apollo}, we adopt the representation scheme of \textsc{Tamme} \cite{tamme} to obtain a uniform token sequence $(\mathbf{e}_t)_{t\in\{1,...,S\}}, \mathbf{e}_t\in\mathbb{R}^d$ from a given multimodal health record: every item in the multimodal health record is considered an individual timestamped event of a very high-level type category from a fixed vocabulary. It may optionally be further specified by free-text type specifics, and it may be associated with a value of a certain modality, such as the contents of an X-ray image, of an ECG, or with a numeric scalar. While the model dynamically learns an embedding for the type categories during pretraining, textual type specifics and values are embedded by non-trainable external encoders. Numeric scalars are represented by Fourier features \cite{fourierfeatures}. All other modalities use the mentioned dedicated external pretrained AI models as encoders. To map all embeddings to a shared space, projection layers are trained end-to-end along with \ours. We prepend every input window with a fixed-length metadata prefix holding one token per demographic attribute (date of birth, race, gender, insurance, language, and marital status). Unlike \textsc{Tamme} \cite{tamme}, we also add dedicated tokens to represent the end of events that are not self-contained but actually go on over a period of time. This includes, in particular, clinical procedures and medication intake. We emphasize that those end tokens are inserted at the timestamp of the physically charted termination of the event, not at an earlier position, to prevent longitudinal information leakage. 
    
\paragraph{Bidirectional temporal attention.} 
    Our multimodal longitudinal patient records are sequences of clinical events distributed across continuous, irregularly spaced intervals. In such trajectories, a discrete ordinal index of an event within the sequence cannot capture the underlying temporal semantics. Standard positional encodings typically either ignore continuous time entirely by relying on ordinal embeddings, whether sinusoidal or learned, or they compress it into a single relative-time axis, such as rotary position embedding (RoPE) \cite{rope}. Neither approach adequately captures a structural property that is fundamental to modeling patient journeys: directional asymmetry of the inter-event temporal gaps. The duration between two adjacent events represents the forward gap of the earlier token and, simultaneously, the backward gap of the later token. When a token acts as an attention \emph{query}, it effectively queries for reachable patient states and the corresponding time frame. Conversely, when acting as a key, it is expected to \emph{implicitly} provide information about the temporal proximity to its predecessor. Encoding both directional views symmetrically makes the model conflate functionally distinct temporal signals. To resolve this, we introduce \emph{bidirectional temporal attention}, an additive temporal encoding that injects a multi-feature temporal position signal directly into the attention mechanism, asymmetric between queries and keys. Let the token at timestep $t \in \{1,\dots,S\}$ in a sequence carry four distinct temporal scalars: an ordinal position index $p_t$, the patient's age $a_t$, the backward time gap since the previous event $\delta^{\mathrm{prev}}_t$, and the forward gap to the next event $\delta^{\mathrm{next}}_t$. These inter-event gaps in our real-world dataset span multiple orders of magnitude. Therefore, we apply a shifted logarithmic transformation to stabilize the dynamic range: $\widetilde\delta^{\,\bullet}_t = \log\bigl(1+\delta^{\,\bullet}_t\bigr)$. We avoid normalizing \emph{across} the temporal features, e.g. via LayerNorm. All features are mapped to a continuous embedding space using a sinusoidal basis. For an $F$-dimensional feature vector $\mathbf{u}=(u_1,...,u_F)\in\mathbb{R}^F$ mapped to a $d_h$-dimensional attention head, we define $N=d_h/(2F)$ frequencies per feature requiring $2F \mid d_h$, geometrically spaced under a base temperature $T=10^4$: 
    \[
    \omega_n = T^{-n/N}, \qquad n=0,\dots,N-1. 
    \]
    Each scalar feature is encoded by its full sine and cosine bank, 
    \[
    \psi(u_i) = \bigl[\sin(u_i\,\omega_0),\dots,\sin(u_i\,\omega_{N-1}),\;\cos(u_i\,\omega_0),\dots,\cos(u_i\,\omega_{N-1})\bigr] \in \mathbb{R}^{2N},
    \]
    and the encodings for each feature are concatenated: 
    \[
    \Psi_F(\mathbf{u}) = \bigl[\,\psi(u_1)\,\Vert\,\cdots\,\Vert\,\psi(u_F)\,\bigr] \in \mathbb{R}^{d_h}.
    \]
    The features shared by queries and keys in the well-known attention computation are position and age $\mathbf{v}^{\mathrm{base}}_t = [p_t, a_t]$. In addition, features used for queries include the time delta to the next event $\mathbf{v}^{\mathrm{fwd}}_t = [\widetilde \delta^{\mathrm{next}}_t]$ while the features for keys include the backward time delta to the previous event $\mathbf{v}^{\mathrm{bwd}}_t = [\widetilde \delta^{\mathrm{prev}}_t]$. During the attention computation, these are combined additively with query ($\mathbf{q}_t$) and key ($\mathbf{k}_t$):
    \[
    \mathbf{q}_t \leftarrow \mathbf{q}_t + \Psi_2(\mathbf{v}^{\mathrm{base}}_t) + \Psi_1(\mathbf{v}^{\mathrm{fwd}}_t), \qquad
    \mathbf{k}_t \leftarrow \mathbf{k}_t + \Psi_2(\mathbf{v}^{\mathrm{base}}_t) + \Psi_1(\mathbf{v}^{\mathrm{bwd}}_t).
    \]
    This decomposition ensures that queries strictly condition on the forward-looking temporal horizon, while keys independently integrate recency. 

\paragraph{Variational autoregression with a context-conditional prior.}
    Standard transformer models in the language domain commonly obtain variability at inference time by sampling from a categorical distribution over a fixed vocabulary. In contrast, our architecture models sequences of composite clinical events, where the output projection comprises multiple component-specific heads predicting continuous embeddings (event value, type specifics) alongside categorical information (event type), timing, etc. Introducing independent variability in each head would decouple the outputs from each other and thus invalidate their overall semantics as parts of one coherent patient state. We therefore route randomness through a single shared continuous latent variable $\mathbf{z}_t\in \mathbb{R}^k$ which is sampled once per timestep $t$ and consumed jointly by all decoder heads. Concretely, we model the conditional distribution of the next event $\mathbf{x}_t$ given the history $\mathbf{x}_{<t} :=(\mathbf{x}_1,...,\mathbf{x}_{t-1})$ via $\mathbf{z}_t$:
    \[
    p_\theta(\mathbf{x}_t \mid \mathbf{x}_{<t}) = \int p_\theta(\mathbf{x}_t \mid \mathbf{z}_t)\, p_\theta(\mathbf{z}_t \mid \mathbf{h}_{t-1})\, d\mathbf{z}_t
    \]
    where $\mathbf{h}_{t-1}$ denotes the transformer output for the patient's record up to $t-1$ under a causal attention mask. The decoder $p_\theta(\mathbf{x}_t \mid \mathbf{z}_t)$ reconstructs the event components of the current event $\mathbf{x}_t$ from $\mathbf{z}_t$. Crucially, the prior $p_\theta(\mathbf{z}_t \mid \mathbf{h}_{t-1})$ is a learned, \emph{context-conditional} diagonal Gaussian, unlike the fixed standard-normal prior of variational autoencoders. Following the framework of sequential latent-variable models \cite{vrnn}, we use a posterior that observes the historical context $\mathbf{h}_{t-1}$ and the input embedding $\mathbf{e}_t$ of the target event $\mathbf{x}_t$:
    \[
    q_\phi(\mathbf{z}_t \mid \mathbf{h}_{t-1}, \mathbf{e}_t) = \mathcal{N}\!\bigl(\boldsymbol{\mu}_q(\mathbf{h}_{t-1}, \mathbf{e}_t),\; \mathrm{diag}\bigl(\boldsymbol{\sigma}_q^2(\mathbf{h}_{t-1}, \mathbf{e}_t)\bigr)\bigr)
    \]
    This is implemented via an MLP that concatenates $\mathbf{h}_{t-1}$ and $\mathbf{e}_t$ and maps them to $(\boldsymbol{\mu}_q, \log \boldsymbol{\sigma}_q)$ via linear heads. The prior conditions strictly on the history:
    \[
    p_\theta(\mathbf{z}_t \mid \mathbf{h}_{t-1}) = \mathcal{N}\!\bigl(\boldsymbol{\mu}_p(\mathbf{h}_{t-1}),\; \mathrm{diag}\bigl(\boldsymbol{\sigma}_p^2(\mathbf{h}_{t-1})\bigr)\bigr)
    \]
    Because causal masking ensures history $\mathbf{h}_{t-1}$ contains no leakage of $\mathbf{x}_t$, any target signal reaches the posterior exclusively through $\mathbf{e}_t$. Training maximizes the per-timestep evidence lower bound (ELBO, see \textbf{Pretraining}) whose regularization term is the KL divergence between posterior and prior. Since both are diagonal Gaussians, it evaluates in closed form \cite{reparametrization-trick-vae}, subscripts $t,j$ indicating the $j$-th component of $\sigma_q, \sigma_p, \mu_q, \mu_p$ at timestep $t$:
    \[
    \mathcal{L}_{\mathrm{KL},t} = \sum_{j=1}^{k} \left[\, \log\!\frac{\sigma_{p,t,j}}{\sigma_{q,t,j}} + \frac{\sigma_{q,t,j}^2 + (\mu_{q,t,j} - \mu_{p,t,j})^2}{2\,\sigma_{p,t,j}^2} - \frac{1}{2} \,\right]
    \]
    Minimizing the KL divergence between posterior and prior forces the model to learn a prior that is close to the posterior. Simultaneously, a reconstruction loss ensures the posterior path learns a bottleneck representation of the current event $\mathbf{x}_t$ like an autoencoder. Together, these two objectives therefore maximize the predictive capability of the prior and thus of the contextual history $\mathbf{h}_{t-1}$. Intuitively, the training objective therefore is to learn an \emph{informative} prior that closely matches an \emph{informed} posterior. To minimize the overall loss, in the long run the posterior is compelled to encode only the unpredictable residual, the \emph{innovation} or \emph{surprise} of the current event. A predictable continuation of the history results in near-zero KL, whereas a surprising update such as acute injury pays a penalty. In our clinical application, a perfect approximation is unlikely, so the remaining prior-posterior gap after training directly imposes an upper limit on inference performance, the genuine unpredictability in clinical reality. To prevent the degenerate solution where the posterior (and prior jointly) collapses to a deterministic prediction ($\boldsymbol{\sigma}_q \to 0$) while passing $\mathbf{e}_t$ through, we apply a sigmoidal squash to the \emph{prior}'s uncertainty:
    \[
    \log \boldsymbol{\sigma}_p \in (\log \sigma_{\min},\, \log \sigma_{\max}), \quad \text{where} \quad \sigma_{\min} = 0.05, \quad \sigma_{\max} = 2
    \]
    With $\boldsymbol{\sigma}_p$ bounded away from zero, $\log(\sigma_p/\sigma_q)$ in the KL divergence increases as $\boldsymbol{\sigma}_q \to 0$. This structurally prohibits the posterior from bypassing the history via perfect certainty. Note that during training, we draw a single sample from the posterior via the reparameterization trick \cite{reparametrization-trick-vae}:
    \[
    \mathbf{z}_t = \boldsymbol{\mu}_q + \boldsymbol{\sigma}_q \odot \boldsymbol{\epsilon}, \quad \boldsymbol{\epsilon} \sim \mathcal{N}(\mathbf{0}, \mathbf{I})
    \]
    Crucially, at inference time with the future target being unavailable, samples are drawn directly from the context-conditional prior:
    \[
    \mathbf{z}_t \sim \mathcal{N}\!\bigl(\boldsymbol{\mu}_p(\mathbf{h}_{t-1}),\; \tau^2\, \mathrm{diag}\bigl(\boldsymbol{\sigma}_p^2(\mathbf{h}_{t-1})\bigr)\bigr)
    \]
    The variance of the clinical trajectories generated this way reflects the model's epistemic uncertainty: a patient trajectory with a definitive, predictable prognosis yields concentrated samples, while a highly ambiguous clinical state naturally results in a broader sampling distribution. Hyperparameter $\tau \in [0, 1]$ controls the degree of this variability. As in transformer-based language models, we therefore refer to $\tau$ as the \emph{temperature}. 

\paragraph{Multimodal output decoding.}
    The latent $\mathbf{z}_t$ must be decoded to reconstruct the multiple components of one clinical event, i.e. type category, value embedding, etc. These components are not independent given $\mathbf{z}_t$. Admissible modalities are constrained by the type category $c_t$. For instance, a medication cannot have an ECG value embedding. Optional free-text type specifics $s_t$ refine the type, and a meaningful value embedding $v_t$ requires that a measurement type and modality $m_t$ have been recovered, too. Independent per-head sampling could therefore produce clinically impossible outputs that invalidate trajectories. We decompose the distribution of components per event along a clinically motivated hierarchy, with $\delta^{\mathrm{next}}_t$ again denoting the temporal gap to the next event:
    \[
    p_\theta(\mathbf{x}_t \mid \mathbf{z}_t) \;=\; p(c_t \mid \mathbf{z}_t)\  p(s_t \mid c_t, \mathbf{z}_t)\  p(m_t \mid c_t, s_t, \mathbf{z}_t)\  p(v_t \mid c_t, s_t, m_t, \mathbf{z}_t)\  p(\delta^{\mathrm{next}}_t \mid \mathbf{z}_t)
    \]
    Cascading feature MLPs mirror the structure of the hierarchy formulated above: first, a type category head extracts features $\mathbf{f}^{\mathrm{cat}}_t$ from the latent $\mathbf{z}_t$ and predicts $c_t$. The subordinate type-specifics head then consumes $\mathbf{z}_t + \mathbf{f}^{\mathrm{cat}}_t$, extracts $\mathbf{f}^{\mathrm{spec}}_t$, and maps to $s_t$. The modality and value embedding heads read $\mathbf{z}_t + \mathbf{f}^{\mathrm{spec}}_t$, and the time head consumes $\mathbf{z}_t$ directly. They retrieve $m_t$, $v_t$, and $\delta^{\mathrm{next}}_t$. The decoded modality $m_t$ selects one of the value path's output heads. The heads for value embedding and type specifics perform regression tasks, aligning reconstructions with the frozen input embedding space. Numeric scalars are emitted as Fourier coefficients in exactly the sinusoidal basis used to encode them in the input sequence. Thus, a sampled numeric token is directly consumable at the next autoregressive step. Inter-event time deltas in our dataset are zero-inflated, since clustered laboratory results, vital sign documentation, and process end markers share similar or identical timestamps. This produces a point mass at $\delta^{\mathrm{next}}_t = 0$ in a heavy-tailed distribution. A single regressor would be pulled toward zero by this mass and systematically underpredict gaps. We therefore split the time head into a binary decision gate $p(\delta^{\mathrm{next}}_t > 0 \mid \mathbf{z}_t)$ that is supervised at every position and a regression head supervised on $\log(1+\delta^{\mathrm{next}}_t)$ only where $\delta^{\mathrm{next}}_t > 0$. The logarithmic transform compresses the tail that spans multiple orders of magnitude. At inference time, we apply a hard gate that, upon a negative decision, clamps the regression output to zero. A soft gate would conflate uncertainty about whether a gap exists with the gap's magnitude. The presence of type specifics $s_t$ is gated identically. The heads for category, modality, and type-specifics presence use data-derived class weights to counteract training data imbalance. Further, we enforce plausible pairings of modality and type category $p(m_t \mid c_t)$ during inference by restricting the Softmax distribution of the modality head to modalities ever observed in the dataset given $c_t$. Overall, the decoder's output scheme is fully compatible with \ours's multimodal input token representation. Thus, a generated token can be re-injected into the model at the next timestep, closing the autoregressive generative loop.

\paragraph{Pretraining.}
    The overall pretraining objective is the negative per-timestep evidence lower bound summed across the patient trajectory and averaged over the training corpus $\mathcal{D}$, augmented by a small auxiliary term for prior reconstruction:
    \[
    \mathcal{L}(\theta, \phi) \;=\; \mathbb{E}_{\mathbf{x}_{1:S}\,\sim\,\mathcal{D}}\!\left[\;     
    \sum_{t=1}^{S} \alpha\,\mathcal{L}_{\mathrm{rec},t}(\mathbf{z}_t) \;+\;               
    \beta\,\mathcal{L}_{\mathrm{KL},t} \;+\;                                              
    \gamma\,\mathcal{L}_{\mathrm{rec},t}(\widetilde{\mathbf{z}}_t) \;\right],   
    \]
    where the latent is drawn either from the informed posterior $\mathbf{z}_t \sim q_\phi(\cdot \mid \mathbf{h}_{t-1}, \mathbf{e}_t)$ or from the context-conditional prior $\widetilde{\mathbf{z}}_t \sim p_\theta(\cdot \mid \mathbf{h}_{t-1})$, both reparameterized \cite{reparametrization-trick-vae}. The prior path mirrors sampling at inference time. The reconstruction loss $\mathcal{L}_{\mathrm{rec},t}$ is the sum of the negative log-likelihoods for all components of the event $\mathbf{x}_t$, i.e. for modality, type category, type specifics, value embedding, and time as described above. Details on the output components are provided above in Section~\textbf{Multimodal output decoding}. The first two terms in the above loss formulation implement the per-timestep ELBO derived in Section~\textbf{Variational autoregression}: $\mathcal{L}_{\mathrm{KL},t}$ is the closed-form Gaussian divergence between posterior and conditional prior, and $\beta$ is annealed linearly from zero to its target value over the initial fraction of training \cite{kl-annealing}. The third term is an auxiliary prior-mode reconstruction loss: at every step the decoder is run a second time on a \emph{prior}-sampled $\widetilde{\mathbf{z}}_t$ and tasked with reconstructing the same $\mathbf{x}_t$ as the posterior path. The coefficient $\gamma$ controls what fraction of the reconstruction budget $\gamma/(\alpha+\gamma)$ is spent on this prior-mode pass. Together with the sigmoidal squash bounding the prior's standard deviation to $(\sigma_{\min}, \sigma_{\max})$, these regularizers structurally exclude two degenerate solutions: first, a deterministic pass-through of the input embedding $\mathbf{e}_t$ with $\boldsymbol{\sigma}_q \to 0$. Second, the prior inflating its variance to evade learning actual predictions. Optimization is done end-to-end with the Muon optimizer \cite{muon} on the hidden 2D weight matrices and AdamW on the remaining parameters under a cosine learning rate schedule with linear warmup. Hyperparameters are reported in Supplementary Table~\ref{tab:hyperparameters}.


\heading{Evaluation framework}

\paragraph{Time-controlled and free forecasting.} We prompt the pretrained \ours model with a patient's history and predict multiple next tokens step by step autoregressively, i.e. each generated token is decoded, re-encoded, and appended to the prompt before the model is applied on the new prompt again for the next timestep. In the plain variant, the model predicts a token and decodes all its event components including type category, type specifics, time, etc. freely (\emph{free forecasting}). In another variant, we enforce a specific time delta $\delta^{\mathrm{next}}_t$ between the current token and the next one (\emph{time-controlled}) through our temporal position integration. We contrast the two variants against each other and evaluate the forecasting capabilities of \ours in general: for each test patient, we set an anchor time~$t^\star$ at a fixed offset of 24 h after their most recent hospital admission, prompt \ours with the patient's history up to $t^\star$, and autoregressively roll out subsequent events for a fixed token budget that aims to cover the held-out window $(t^\star, t^\star{+}H_{\max}]$ up to the maximum horizon $H_{\max}$. Because each per-timestep latent $\mathbf{z}_t$ is sampled fresh from the context-conditional prior $p_\theta(\mathbf{z}_t \mid \mathbf{h}_{<t})$ at every generated event and then decoded by $p_\theta(x_t \mid \mathbf{z}_t)$, our $M{=}50$ independent rollouts realize a Monte Carlo (MC) approximation of $p_\theta(x_{>t^\star} \mid x_{\leq t^\star})$. The empirical MC frequency then gives the probability of event type category $c$ or modality $m$ occurring within $(t^\star, t^\star{+}H]$. We evaluate these predicted probabilities against the held-out ground truth across horizons $H \in \{1, 2, 4, 6, 12, 24, 48, 72\}$ hours. Contrasting free and time-controlled performance then isolates the share of forecasting accuracy attributable to correctly anticipating event \emph{timing} versus event \emph{content}, per class and per horizon. We report macro AUROC, macro Brier, and a macro count error, i.e. predicted versus actual counts per class, for high-level type category and modality. In the absence of a model directly comparable to \ours, we benchmark against a prevalence chance floor and a trivial persistence baseline that labels class $c$ or $m$ as recurring in $(t^\star, t^\star{+}H]$ if and only if it occurred in the pre-anchor window $(t^\star{-}H, t^\star]$.

\paragraph{Zero-shot Monte Carlo classification.} The rollout primitive and Monte Carlo estimation described above enable estimating the probability of certain clinical target events within a specific future time frame. By this, \ours supports training-free classification for standard clinical decision problems, thus zero-shot. We evaluate binary classification for prolonged length of stay (LOS) exceeding 72 h, mortality within 72 h, and 30-day readmission. A reference time $t_\mathrm{ref}$ defines the clock origin. For the length of stay and the mortality tasks, $t_\mathrm{ref}$ is the time of admission, while for the readmission task $t_\mathrm{ref}$ is the time of discharge. The model is prompted with the patient records up to rollout onset $t^\star \geq t_\mathrm{ref}$. We set $t^\star = t_\mathrm{ref} + 48$ h for LOS and mortality, while for readmission we set $t^\star = t_\mathrm{ref}$, yet truncate the prompt for readmission at one timestep before $t^\star$ to avoid label leakage by inter-event time deltas. Per patient we draw $M$ trajectories from $\mathbf{x}_{\le t^\star}$ via \ours's autoregressive rollout \textit{without} time control. For LOS and mortality, we use $M=50$, while for readmission we reduce to $M=25$ due to a longer forecasting horizon necessitating an increased token budget and computation time. For each rollout, we classify by binning the time of the first generated target event relative to $t_\mathrm{ref}$ against a task-specific threshold: we evaluate for a prolonged stay longer than 72 h from admission to discharge, death within 72 h of admission, and readmission within 30 d of discharge. A trajectory that produces no target token but whose simulated horizon reaches past the threshold is assigned to the second bin. Thus, it is counted as a prolonged stay for LOS and negative for mortality and readmission. A simulation that exhausts the generation token budget before either condition is satisfied is declared \emph{invalid} and excluded. We compute the predicted probability as the frequency over the valid trajectories. To prevent degenerate results under a low number of rollouts and to reduce the influence of patients with few valid rollouts, we apply Laplace smoothing with an additive constant of $0.5$. We report coverage, i.e. the mean fraction of valid trajectories per patient, and conversely the invalid-run ratio alongside the AUROC, balanced accuracy, Brier score, and ECE. The first generated event is seeded with a task-specific forward time delta $\delta^{\mathrm{next}}_t$ from the prompt's last visible token to avoid the rollout stacking at the initial timepoint. To reduce computational cost, we evaluate on 2000 randomly selected test set patients for LOS and mortality, and on 1000 patients for the more resource-intensive task of readmission prediction.

\paragraph{Counterfactuals.} To evaluate \ours's capability of counterfactual intervention simulation, we choose a problem of high relevance that allows for direct empirical comparison: we formulate \ours's counterfactual evaluation as an emulation of the SMART trial~\cite{counterfactual-medical-study}: for early fluid resuscitation in critically ill adults, it compared 0.9\% saline with balanced crystalloids, predominantly lactated Ringer's solution as two different possible interventions. Fluid resuscitation is a cornerstone of early sepsis management, yet the optimal crystalloid remains debated since saline has been associated with hyperchloremia, metabolic acidosis, renal vasoconstriction, and acute kidney injury, whereas balanced crystalloids more closely resemble physiological plasma composition. In SMART, balanced crystalloids reduced the incidence of major adverse kidney events within 30 days (MAKE-30: occurrence of at least one of new renal-replacement therapy (RRT), death, or persistent renal dysfunction within 30 d), with particularly pronounced benefit observed among patients with sepsis, making fluid choice a clinically meaningful intervention for counterfactual evaluation \cite{counterfactual-medical-study, counterfactual-medical-study-sepsis, evans2021surviving}. These findings have raised the important question of how alternative fluid strategies may affect outcomes in individual patients. In our setting, we first identify patients with sepsis markers from \ours's test set. We then substitute the respective fluid in their factual multimodal event sequences and from the intervention onward perform autoregressive rollout in the Monte Carlo fashion described above to estimate outcome probabilities under different actions. We investigate whether \ours's simulation yields results similar to the sepsis cohort in the clinical trial SMART \cite{counterfactual-medical-study-sepsis}. We identify patients with sepsis markers and administration of either 0.9\% NaCl or lactated Ringer's within six hours of the marker. The first administration time of either fluid within that time window defines $t^\star$. We exclude patients whose records end before $t^\star + 24$ h without a terminal event, i.e. without death or discharge, as well as patients whose entire record up to $t^\star + 24$ h comprises fewer than 256 events in total. For patients with multiple eligible episodes, we include only the chronologically earliest, to avoid information leakage by trends or patterns in a patient's history. This results in $1{,}085$ samples in total. Per sample, we prompt \ours with the patient's lifetime trajectory up to $t^\star$ and simulate future development via autoregressive rollout under two arms: (1) the \emph{factual} arm that receives the fluid actually administered and thus uses the prompt without changes, and (2) the \emph{counterfactual} arm where the type of the fluid actually received is substituted with the type of the other one. We preserve the time of the factual event, so the contrast isolates the effect of fluid composition. For each arm we draw $M=25$ trajectories via the same free-mode autoregressive sampling used above. During this rollout, we constrain any generated crystalloid administration to match the type of the fluid associated with the current arm. Further, we again seed the first generated event per rollout with a forward time delta $\delta^{\mathrm{next}}_t$ of 1 h. Both arms share random numbers and receive identical latent noise. We score the outcomes and again perform Monte Carlo estimation to obtain their empirical probability: $30$-day in-hospital mortality, length of stay, and the composite MAKE-30 score, all anchored at $t^\star$. Computation and filtering of invalid rollouts are handled just as above in Section~\textbf{Zero-shot Monte Carlo classification}. To quantify the average treatment effect, i.e. the difference in the effects each of the fluids has on patients when used as sepsis interventions, we collect the outcome probabilities for mortality, LOS, and MAKE-30 and measure the difference between the two arms. We then average over all patients. For investigating simulated patient state trajectories qualitatively, we apply chain-UMAP projection to the generated patient states of each arm and also a trained MLP to infer the NEWS2 risk score from each of them (see Section \textbf{NEWS2 scoring}).

\paragraph{Linear probing.} Under causal attention, \ours's contextualized transformer output $\mathbf{h}_t$ before the prior network is a representation of the patient state after having observed events for timesteps $0,\dots,t$. We treat it as the \emph{patient state embedding} at time $t$ and write $\mathbf{h}_\mathrm{final}$ for its value at the last observed event of a patient's record. For every test patient, we extract all $\mathbf{h}_t$ along the whole observed trajectory. We then apply linear probing for different timepoints within and around the last stay in a patient's records. This can be an ICU or general hospital stay depending on the task we probe for. A probe is an $\ell_2$-regularized logistic regression ($C{=}1$, L-BFGS) on inputs that are standardized per feature. We estimate performance by stratified five-fold cross-validation repeated over $3$ random seeds, resulting in $15$ independent fits per target and position. We report mean $\pm$ s.d.\ of AUROC across folds and accuracy as secondary metric. The following families of tasks are evaluated for each patient: first, clinical outcomes including prolonged hospital and ICU stay ($\geq 72$ h), early in-hospital mortality ($\leq 72$ h after admission), and overall in-hospital mortality (death at any time during the stay). Second, concept recovery in one-vs-rest probes for each of $15$ high-level ICD chapters and $29$ Quan-Elixhauser comorbidity flags~\cite{quan2005coding}. A target is probed at a given position only where each class has at least five instances per fold. To trace the inclusion of clinical information in the patient embedding as a stay unfolds, every probe is re-fit at positions defined by absolute time since hospital admission: pre-admission, admission, $+12$, $+24$, $+36$, and $+48$ h, and discharge ("leave"). A position is dropped for patients whose stay ends before the respective horizon. Thus, the number of samples changes per position and is annotated in our results. The embedding at the "leave" position has access to all information of the stay, hence probing it provides clean insights about the embeddings' capabilities for \emph{retrieval}. For all other positions, the amount and kind of information available in each position vary between stays. Thus, we consider all positions before "leave" as read-outs that mix prediction and retrieval, and conservatively evaluate them as retrieval.

\paragraph{NEWS2 scoring.} To probe for time-varying clinical risk indicators, we train and evaluate an MLP to retrieve the National Early Warning Score 2 (NEWS2)~\cite{news2} from single \ours patient state embeddings $\mathbf{h}_t$. Ground truth is computed deterministically from the charted vital measurements in the patients' event streams of \ours's test set. From this set, we hold out randomly selected patients for evaluating the MLP. \ours's training data never enters this MLP. For ground truth computation, a scoring window opens at the first of the five vitals required for computing NEWS2 (respiration rate, oxygen saturation, temperature, systolic blood pressure, heart rate), keeps the latest value per parameter, and closes once all five are present. It is capped at a total window span of one hour. Windows are non-overlapping and never span across patients. The more sparsely charted oxygen-supplementation and consciousness observations (Level-of-Consciousness text, otherwise mapped from the Glasgow Coma Scale) do not open windows but are carried forward for up to eight hours. We default to "breathing room air" and "being alert". Items are matched by exact identifiers with unit conversion, and implausible values are dropped. Only ICU and emergency department settings chart all five required vitals jointly. Overall, we obtain $1.83$ million windows, $272{,}698$ of them over the \ours test patients. Each window is anchored at the token of its last contributing event. We train and probe with the patient state $\mathbf{h}_t$ at that anchor associated with the NEWS2 score computed from that window. At this point, the embedding has observed exactly the vitals the label is computed from, making this a \emph{retrieval} probe of the current state, not a forecast of deterioration. Notably, in other evaluation settings such as counterfactual intervention simulation, we apply the trained MLP not only to embeddings anchored on a ground truth vital window, but to \emph{any} patient state embedding at \emph{any} point in time, including the states generated during autoregressive rollout. For training and testing the MLP we split patients by a 70/15/15 ratio, standardize features with training set statistics, and fit an MLP with hidden layers of sizes 512 and 256 (LayerNorm, GELU, dropout $0.1$) by minimizing the MSE with AdamW. We use a learning rate of $10^{-3}$, weight decay $0.01$, batch size $4096$, at most $60$ epochs, and early stopping on the validation loss. Three seeds yield near-identical results. We report MAE, RMSE, $R^2$, Spearman correlation, quadratic-weighted $\kappa$ over the three commonly established NEWS2 risk bands, and AUROC at the NEWS2 $\geq 5$ and $\geq 7$ levels, with confidence intervals from resampling patients ($1{,}000$ bootstrap draws). We additionally report the within-patient $R^2$, which scores against each patient's own label variance. Also, we contrast against trivial constant prediction, linear-regression, and age-only baselines. Finally, as the trained MLP maps \emph{any} \ours patient state embedding to a NEWS2 score, we apply the trained model unchanged to the generated states of autoregressive rollouts in counterfactual intervention simulation to track the simulated risk over time (see Fig.~\ref{fig:4}).

\paragraph{Survival analysis.} Time-to-event or survival analysis~\cite{kleinbaum12} models the data as a set of $\{x_i, t_i, e_i\}$, with $x_i\in\mathbb{R}^d$ the $d$-dimensional embedding of the patient stay, $e_i \in \{0, 1\}$ indicates the occurrence of event ($e_i=1$, death after admission) or right-censoring ($e_i=0$, discharge without death), $t_i \in \mathbb{R}^+$ the time until event or censoring for the subject $i$. The prediction targets are the survival function $S(t|x)$, which indicates the probability that an individual survives beyond time $t$, and the hazard function $h(t|x)$, which describes the instantaneous rate of event conditioned on surviving up to time $t$:
\[
    S(t|x) = \mathbb{P}(T>t|X=x) = e^{-\int_0^th(u|x)du}
\]
We choose DeepSurv~\cite{katzman18}, the deep learning extension of the Cox Proportional Hazard model~\cite{cox72}, to effectively incorporate the high-dimensional embeddings~\cite{huo25}.  This assumes the hazard function is expressed as a baseline hazard $h_0(t)$ and the risk score $h(x)$, predicted by a neural network:
\[
    h(t|x) = h_0(t)e^{h(x)}
\]
To evaluate predictive performance changes with increasing longitudinal context in embeddings, we build six prediction settings indexed by a within-stay time ($T$) fraction ($p \in \{0.0, 0.2, 0.4, 0.6, 0.8, 1.0\}$, where $p=0.0$ is entering the hospital, and $p=1.0$ is the last event before death or discharge). For this, we retain only patients with more than five recorded events and at least one hour between hospital admission and discharge ($N{=}26{,}476$). We select the last clinical event with a stay-relative time strictly below the time fraction, resolving concurrent events at the same timestamp by taking the last event.

For each of six prediction settings, we employ nested 5-fold cross-validation, with 10\% of each training set reserved for hyperparameter tuning and an additional 10\% for early stopping, as in~\cite{jeanselme23}. We evaluate the discriminative performance using the time-dependent C-index, $C^{td}$ ~\cite{antolini05}. For hyperparameter optimization, we use the Tree-Structured Parzen Estimator (Bayesian optimization) in Optuna~\cite{optuna} with 100 trials, setting $C^{td}$ as the optimization target. For the hyperparameter optimization space, see Supplementary Table~\ref{tab:hyparams:tte}. Additionally, we assess the calibration performance using D-Calibration (D-Cal)~\cite{haider20}, visualizing as in~\cite{sabook} and providing p-values for each fold in Supplementary Table~\ref{tab:deepsurv_results}. The numerical values for $C^{td}$ and D-cal, as well as the IBS as a joint metric of calibration and discrimination, are reported in Supplementary Table~\ref{tab:deepsurv_results}.

To assess marginal calibration, we use \textit{Distribution Calibration} (D-Cal)~\cite{haider20}, which shows how well predicted survival probabilities align with observed outcomes based on a goodness-of-fit test. D-Cal discretizes the predicted survival probabilities at the true event times into $n$ equidistant intervals in $[0,1]$, and performs a chi-squared test for the uniformity of the distribution. As suggested in the original paper~\cite{haider20}, we set $n=20$ and report the p-values for each fold, with p-values $< 0.05$ indicating miscalibrated predictions. 

 
\newpage
\heading{Computing environment}
\ For all model development, training, experiments, and analysis, we use PyTorch 2.10 on Python 3.12 with CUDA 12.8. Everything can be fully replicated using open-source libraries. We train and evaluate on up to 8$\times$ NVIDIA A100 GPUs (80 GB).

\heading{Data and code availability}
\ All data are available from the original providers and distributors of the Medical Information Mart for Intensive Care (MIMIC) dataset family \cite{mimiciv, mimic-note, mimiccxr, mimiccxr-orig, mimiced, mimic-ecg, mimic-echo} via PhysioNet \cite{physionet} and must not be redistributed by us. Our code will be provided on GitHub for academic research purposes upon publication.

\heading{Author contributions}
\ T.S. conceived the study and developed the model. T.S. and R.R. prepared the manuscript. T.S., Ö.T., and M.E.L. curated the dataset and performed data preprocessing. Ö.T. contributed to model pretraining. T.S., D.S., and R.R. designed the evaluation framework and carried out the experiments and analysis. R.R., L.S., R.B., and D.R. provided critical insights and conceptual guidance. L.S. and R.B. provided medical expertise and clinical analysis. R.B. and D.R. supervised the research. All authors contributed to the writing. All authors reviewed and approved the final manuscript. 

\heading{Acknowledgments}
\ This work benefited from resources provided by the joint project "Open Medical Inference", a Module 3 project of the Medical Informatics Initiative of the Federal Government of Germany, funded by the German Federal Ministry for Research, Technology, and Space (Grant Number 01ZZ2315B). 
 
\end{spacing}

\newpage
\begin{nolinenumbers}
\heading{References} 
\vspace{2mm}
\begin{spacing}{0.9}
\bibliographystyle{naturemag}
\bibliography{literature}
\end{spacing}
\end{nolinenumbers}

\clearpage
\setcounter{figure}{0}
\setcounter{table}{0}
\renewcommand{\figurename}{\textbf{Extended Data Figure}}
\renewcommand{\tablename}{\textbf{Extended Data Table}}
\begin{nolinenumbers} \newpage
\heading{Extended Data Figures}

\begin{figure}[ht!]
    \centering
    \includegraphics[width=\linewidth]{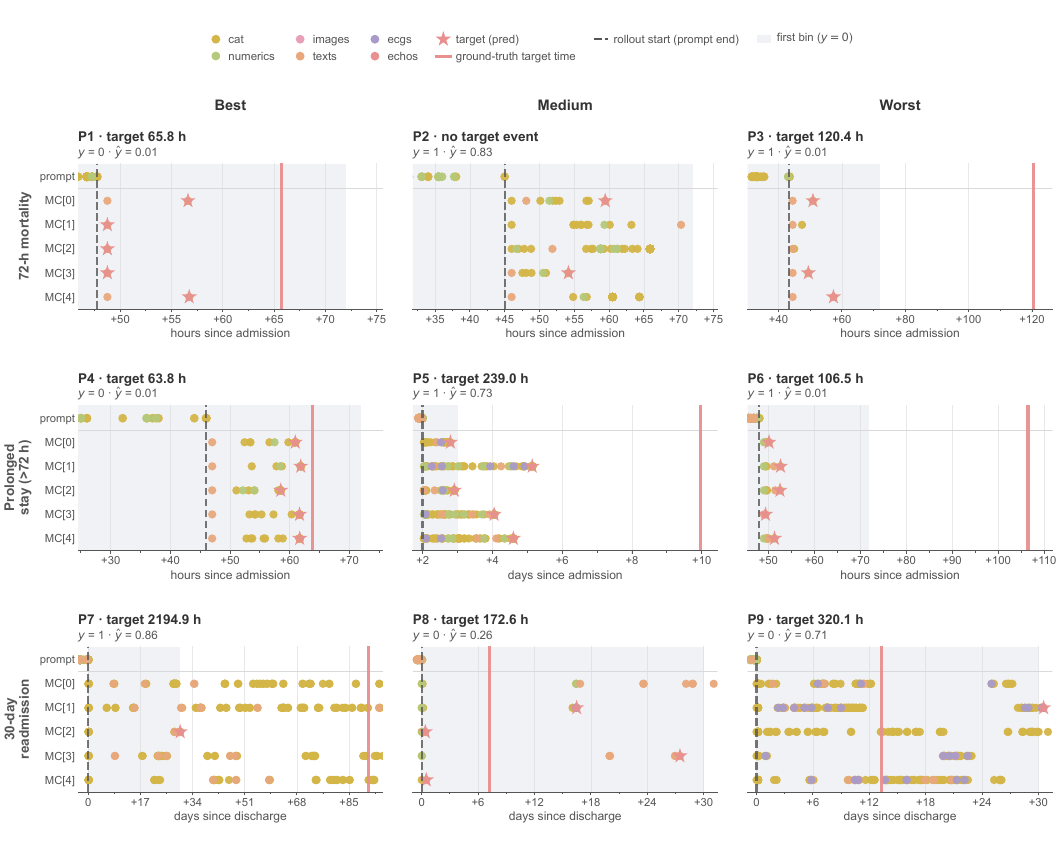}
    \caption{\textbf{Zero-shot classification examples.} Examples of Monte Carlo rollouts for 72 h mortality, prolonged hospital stay ($>$ 72 h), and 30-day readmission. Each with ground truth binary label $y$ and predicted probability $\hat{y}$. The model sees the patient's history and current events up to a certain cutoff (dotted line). We then roll out the next tokens within a fixed budget until the target event (star) is predicted, the upper edge of the class 0 bin (gray area) is reached, or the token budget is exhausted. The red line marks the position of the ground truth event, either within class 0 or not.}
    \label{fig:res_zs_ex}
\end{figure}

\clearpage
\begin{figure}[ht!]
    \centering
    \includegraphics[width=\linewidth]{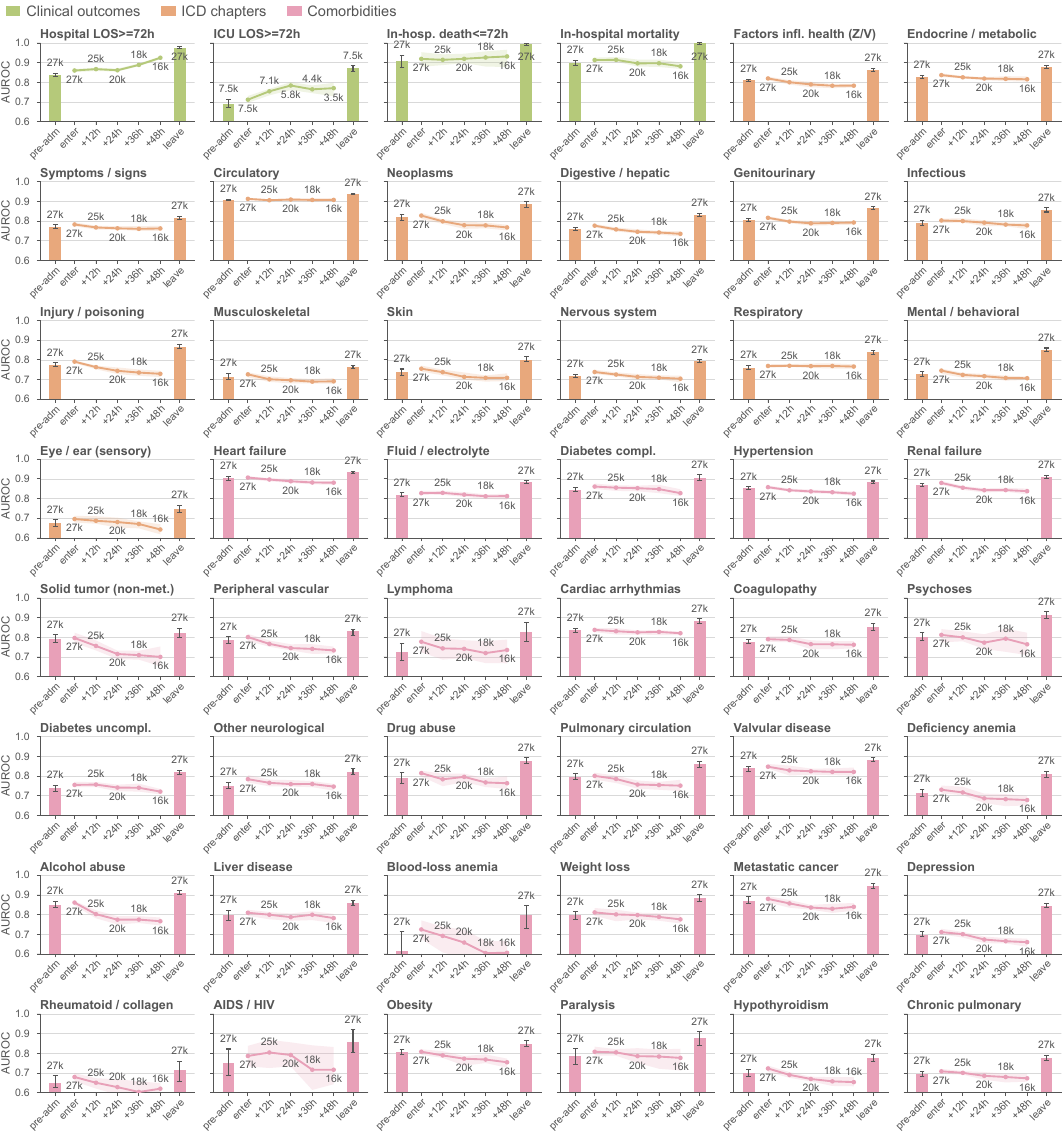}
    \caption{\textbf{Linear probing of \ours's patient states over time.} Cross-validated AUROC of linear probes fit on the frozen patient state $h$ at successive points of each test patient's last stay. Each ICD and comorbidity panel is a binary classification with ground truth collected at discharge. At the pre-admission probe, the model has seen only the patient's history, but no data after the formally charted beginning of the current stay. The probe at "leave" is a clean retrieval. Earlier positions are read-outs mixing prediction and retrieval, since they may carry information on ICD chapters and comorbidities after some timepoint. Probes are $\ell_2$-regularized logistic regression, one fit per position. Plots show the mean and 2.5-97.5th percentile AUROC across all folds. Bar annotations show the number of samples per timepoint.}
    \label{fig:res_lp_tmp}
\end{figure}

\begin{figure}[ht!]
    \centering
    \includegraphics[width=\linewidth]{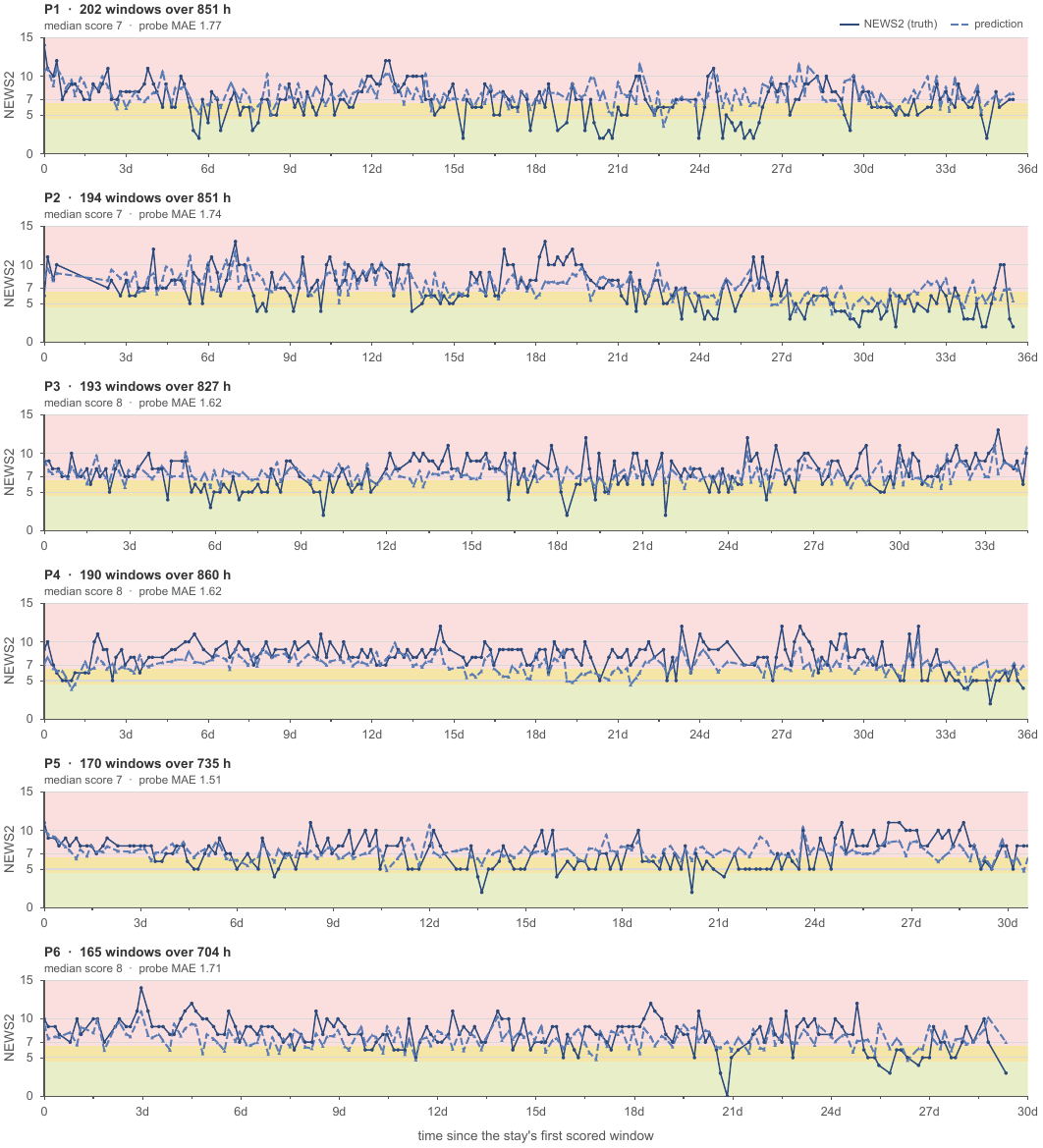}
    \caption{\textbf{Read-out of NEWS2 risk score from \ours's patient states over time.} On our test set, we build an MLP to map patient state embeddings to the NEWS2 risk score. For this, we compute ground truth deterministically on windows where the necessary vitals are all available. We then evaluate on a held-out set and obtain the exemplar predictions above, one time series per patient. The model overall performs very well in distinguishing the widely established risk zones of NEWS2 (green to red shading). It only deviates significantly \emph{within} the high-risk area (red zone).} \label{fig:res_lp_risk}
\end{figure}

\begin{figure}[ht!]
    \centering
    \includegraphics[width=\linewidth]{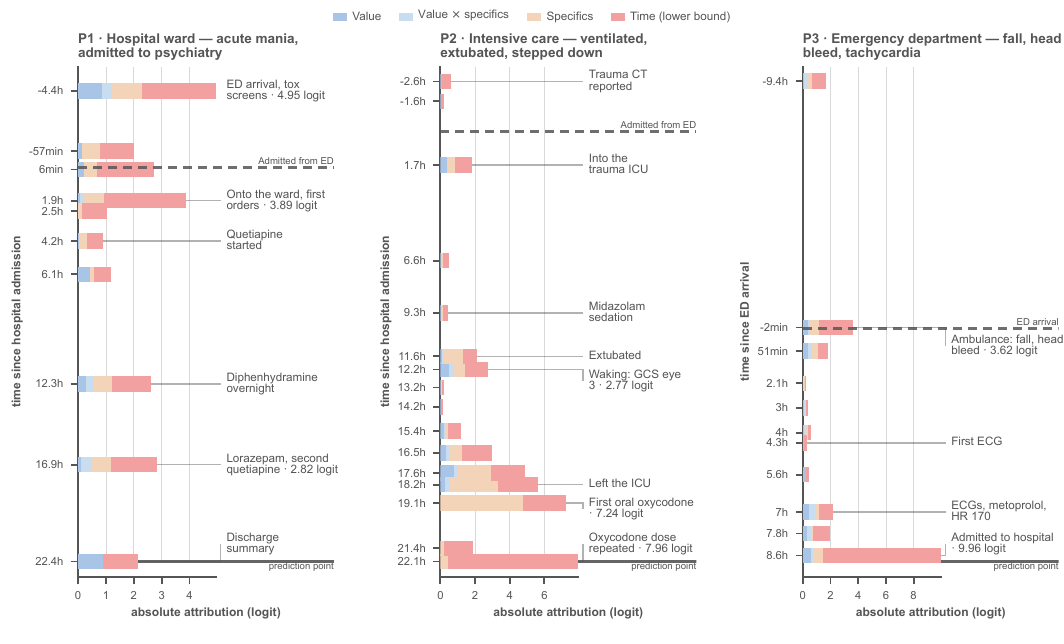}
    \caption{\textbf{\ours's attribution over exemplar patient timelines.} We visualize how much each input token and its components (type category, specifics, time, and value) contribute to the prediction of the event type of the next token following these visualized prompts. Overall, time plays the most crucial role. Value is most important in events where the type alone is actually not meaningful in a clinical sense, e.g. lab results for tox screening, discharge summary, and ECG, while time and specifics usually dominate. Attribution values are computed using Integrated Gradients with an "event type"-only baseline.}
    \label{fig:res_ip_ex}
\end{figure}

\begin{figure}[ht!]
    \centering
    \includegraphics[width=\linewidth]{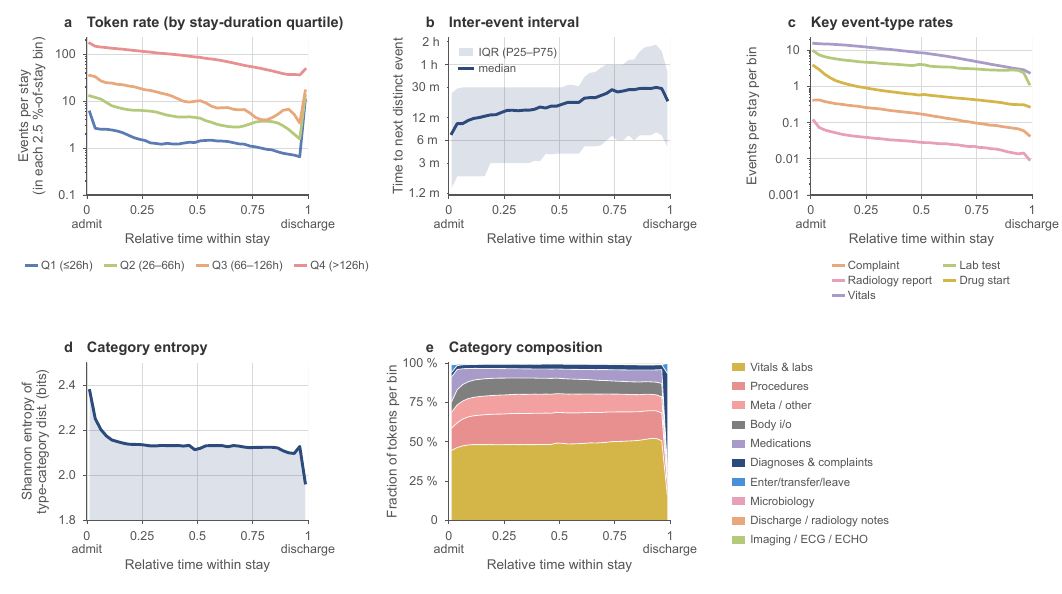}
    \caption{\textbf{Within-stay distribution of events in our dataset.} Events are binned by relative temporal progress within a hospital stay ($0=$ admission, $1=$ discharge). Each stay contributes once (all stays $\geq 1$ h). (a) Mean number of events per stay per bin by stay-duration quartile. The rate of collected events is highest at the beginning of a stay, decreases steadily over the progression of the stay, and rebounds sharply in the final bin. (b) Median time to the next distinct event. Gaps lengthen from ${\sim}7$ to ${\sim}30$ minutes over the first three quarters of the stay and plateau toward discharge. (c) Rates of key event types. (d) Shannon entropy of the type category distribution. Informational variety peaks at admission, plateaus mid-stay, and drops toward the end. (e) Average composition of event type categories over time.}
    \label{fig:ds_instay}
\end{figure}
 \newpage
\clearpage
\heading{Extended Data Tables}

\begin{table}[ht!]
    \centering
    \caption{\textbf{Forecasting.} Generative forecasting of future clinical events. Given each patient record up to a fixed cutoff (stay-anchored, prompt cut at enter time $+24$ h), \ours draws 50 Monte Carlo rollouts and we score whether each event-type ($C{=}31$) or modality ($C{=}6$) class actually occurs within a horizon $H$. Macro-averaged AUROC and Brier score at four horizons, $N{=}13{,}551$ patients. \ours (free) predicts both event content and timing while the time-controlled variant is provided the ground truth forward time delta at each next token prediction. Persistence repeats the previous $H$-window looking in the opposite temporal direction, i.e. on the prompt. Marginal is the prevalence floor (chance). Best model per column is in \textbf{bold}. Parentheses give 95\% patient-bootstrap confidence intervals ($B{=}1{,}000$ resamples).}
    \label{tab:forecasting-headline}
    \setlength{\tabcolsep}{4pt}
    \resizebox{\textwidth}{!}{
    \begin{tabular}{llrrrr}
        \toprule
        {\bf Predictor} & {\bf Granularity} & \bf 1 h & \bf 6 h & \bf 24 h & \bf 72 h \\
        \midrule
        \multicolumn{6}{l}{\textit{Macro AUROC}\quad{\small $\uparrow$ better}} \\
        \addlinespace[4pt]
        Marginal & event-type & 0.500 (0.500, 0.500) & 0.500 (0.500, 0.500) & 0.500 (0.500, 0.500) & 0.500 (0.500, 0.500) \\
        Persistence & event-type & 0.582 (0.580, 0.593) & 0.627 (0.624, 0.630) & 0.610 (0.604, 0.618) & 0.592 (0.585, 0.598) \\
        \ours (free) & event-type & 0.747 (0.714, 0.756) & 0.826 (0.810, 0.847) & 0.829 (0.818, 0.840) & 0.820 (0.813, 0.827) \\
        \ours (time-controlled) & event-type & \textbf{0.866} (0.853, 0.903) & \textbf{0.854} (0.847, 0.861) & \textbf{0.856} (0.846, 0.866) & \textbf{0.842} (0.835, 0.849) \\
        \addlinespace[3pt]
        Marginal & modality & 0.500 (0.500, 0.500) & 0.500 (0.500, 0.500) & 0.500 (0.500, 0.500) & 0.500 (0.500, 0.500) \\
        Persistence & modality & 0.657 (0.652, 0.693) & 0.604 (0.597, 0.611) & 0.583 (0.576, 0.592) & 0.571 (0.565, 0.579) \\
        \ours (free) & modality & 0.764 (0.702, 0.779) & 0.818 (0.790, 0.838) & 0.811 (0.799, 0.822) & 0.813 (0.802, 0.824) \\
        \ours (time-controlled) & modality & \textbf{0.952} (0.934, 0.960) & \textbf{0.894} (0.868, 0.919) & \textbf{0.839} (0.827, 0.852) & \textbf{0.828} (0.816, 0.839) \\
        \midrule
        \multicolumn{6}{l}{\textit{Macro Brier score}\quad{\small $\downarrow$ better}} \\
        \addlinespace[4pt]
        Marginal & event-type & 0.050 (0.050, 0.050) & 0.088 (0.088, 0.088) & 0.106 (0.106, 0.106) & 0.120 (0.120, 0.120) \\
        Persistence & event-type & 0.059 (0.058, 0.061) & 0.119 (0.117, 0.120) & 0.274 (0.272, 0.276) & 0.377 (0.375, 0.379) \\
        \ours (free) & event-type & 0.045 (0.044, 0.046) & 0.073 (0.072, 0.074) & 0.092 (0.090, 0.093) & \textbf{0.105} (0.104, 0.107) \\
        \ours (time-controlled) & event-type & \textbf{0.032} (0.032, 0.033) & \textbf{0.061} (0.060, 0.063) & \textbf{0.089} (0.088, 0.090) & 0.110 (0.109, 0.111) \\
        \addlinespace[3pt]
        Marginal & modality & 0.111 (0.111, 0.111) & 0.116 (0.116, 0.116) & 0.102 (0.102, 0.102) & 0.097 (0.097, 0.097) \\
        Persistence & modality & 0.091 (0.089, 0.093) & 0.147 (0.145, 0.150) & 0.147 (0.145, 0.149) & 0.174 (0.171, 0.176) \\
        \ours (free) & modality & 0.097 (0.095, 0.098) & 0.101 (0.100, 0.103) & 0.094 (0.093, 0.096) & 0.087 (0.086, 0.089) \\
        \ours (time-controlled) & modality & \textbf{0.038} (0.037, 0.040) & \textbf{0.075} (0.073, 0.077) & \textbf{0.092} (0.091, 0.094) & \textbf{0.084} (0.082, 0.085) \\
        \bottomrule
    \end{tabular}
    }
\end{table}

\begin{table}[ht!]
    \centering
    \caption{\textbf{Zero-shot Monte Carlo estimation.} \ours estimates the empirical probability of each clinical outcome via occurrences of the target event across Monte Carlo rollouts without task-specific training (zero-shot). Cells are point estimates with 95\% patient-level bootstrap confidence intervals. Coverage is the fraction of autoregressive rollouts producing a scorable outcome. The lower half provides details on the configuration for each task.}
    \label{tab:zeroshot}
    \begin{tabular}{lccc}
        \toprule
         & {\bf Prolonged stay ($>$ 72 h)} & {\bf 30-day readmission} & {\bf 72 h mortality} \\
        \midrule
        AUROC $\uparrow$ & 0.741 (0.710, 0.771) & 0.609 (0.560, 0.655) & 0.972 (0.943, 0.996) \\
        Balanced accuracy $\uparrow$ & 0.683 (0.652, 0.714) & 0.569 (0.534, 0.603) & 0.821 (0.588, 0.990) \\
        Brier $\downarrow$ & 0.203 (0.191, 0.216) & 0.232 (0.211, 0.252) & 0.035 (0.028, 0.042) \\
        ECE $\downarrow$ & 0.201 & 0.146 & 0.085 \\
        \cmidrule(lr){1-4}
        Coverage $\uparrow$ & 0.804 & 0.645 & 0.822 \\
        \cmidrule(lr){1-4}
        Prediction time & admission $+$48 h & discharge & admission $+$48 h \\
        MC rollouts $M$ & 50 & 25 & 50 \\
        Rollout length & 128 tokens & 1,024 tokens & 128 tokens \\
        Temperature & 1 & 1 & 1 \\
        \bottomrule
    \end{tabular}
\end{table}

\begin{table}[ht!]
    \centering
    \caption{{\bf Counterfactual simulation.} Simulated effect of saline versus balanced crystalloid (Ringer's) on 30-day in-hospital mortality and length of stay ($N{=}1{,}085$ patients). For each patient we run simulations under two arms, the runs sharing the same prompt up to the factual versus the counterfactual fluid given at the real treatment time. \emph{Matched} is the result for the fluid actually received. Parentheses give patient-bootstrap 95\% intervals ($B{=}1{,}000$). Paired contrasts use only patients with a defined outcome in \emph{both} arms, while each scenario row uses all patients defined in \emph{its} arm. In particular, Effect is not the difference of the rows above for LOS, where a simulation without discharge leads to an undefined outcome. The sepsis subgroup of the SMART trial serves as reference: 30-day in-hospital mortality $31.2$ vs. $26.3$\% for saline vs. balanced crystalloid, adjusted OR $0.74$ ($95$\% CI $0.59$-$0.93$) \cite{counterfactual-medical-study-sepsis}. The interval shown is derived from the published arm rates (unadjusted).}
    \label{tab:counterfactual-headline}
    \setlength{\tabcolsep}{6pt}
    \begin{tabular}{lccc}
        \toprule & \multicolumn{2}{c}{\bf 30-day in-hospital mortality (\%)} & {\bf Hospital LOS (h)} \\
        \cmidrule(lr){2-3} \cmidrule(lr){4-4}
        {\bf Scenario} & {\it over runs with explicit outcome} & {\it over all runs} & \\
        \midrule
        Observed & \multicolumn{2}{c}{$21.0$ ($18.9$, $23.5$)} & $253.7$ ($238.8$, $270.4$) \\
        Matched {\footnotesize(calibration)} & $73.7$ ($72.0$, $75.4$) & $48.4$ ($46.9$, $50.2$) & $170.1$ ($163.5$, $176.9$) \\
        Saline & $74.6$ ($72.9$, $76.4$) & $49.2$ ($47.6$, $50.9$) & $174.3$ ($168.2$, $181.4$) \\
        Ringer's & $65.8$ ($63.8$, $67.7$) & $44.9$ ($43.2$, $46.7$) & $135.7$ ($131.4$, $140.1$) \\
        \midrule
        {\bf Effect} {\footnotesize(Saline $-$ Ringer's)} & $+8.86$ pp ($7.83$, $9.89$) & $+4.30$ pp ($3.29$, $5.32$) & $+40.5$ ($35.3$, $46.0$) \\
        SMART, sepsis \cite{counterfactual-medical-study-sepsis} & $+4.9$ pp ($+0.4$, $+9.2$) & - & - \\
        SMART, all \cite{counterfactual-medical-study} & $+0.8$ pp (n.s.) & - & - \\
        \bottomrule
    \end{tabular}
\end{table}

\begin{table}[ht!]
  \centering
  \small
  \setlength{\tabcolsep}{4pt}
  \renewcommand{\arraystretch}{0.95}
  \caption{\textbf{Dataset.} Cohort demographic and clinical characteristics over all patients. Admission-derived attributes (age at first admission, length of stay, in-hospital mortality) are defined only for hospitalized patients. All percentages are of the full cohort.}
  \label{tab:dataset-cohort-demographics}
  \begin{tabular}{@{}llr@{}}
    \toprule
    \textbf{Attribute} & \textbf{Value} & \textbf{Count} \\
    \midrule
    Hospitalized & Yes & 180,733 (60.3\%) \\
    & No & 118,979 (39.7\%) \\
    \midrule
    Sex & Female & 158,525 (52.9\%) \\
    & Male & 141,187 (47.1\%) \\
    \midrule
    Age at first admission (years) & 18-29 & 24,596 (8.2\%) \\
    & 30-44 & 31,277 (10.4\%) \\
    & 45-64 & 58,410 (19.5\%) \\
    & 65-79 & 41,010 (13.7\%) \\
    & 80+ & 25,440 (8.5\%) \\
    \midrule
    Race or ethnicity & White & 166,855 (55.7\%) \\
    & Black or African American & 36,941 (12.3\%) \\
    & Hispanic or Latino & 16,010 (5.3\%) \\
    & Asian & 13,370 (4.5\%) \\
    & American Indian or Alaska Native & 558 (0.2\%) \\
    & Native Hawaiian or other Pacific Islander & 341 (0.1\%) \\
    & Other & 15,202 (5.1\%) \\
    & Unknown & 50,435 (16.8\%) \\
    \midrule
    Insurance & Unknown & 118,979 (39.7\%) \\
    & Other & 108,817 (36.3\%) \\
    & Medicare & 57,735 (19.3\%) \\
    & Medicaid & 14,181 (4.7\%) \\
    \midrule
    Hospitalizations per patient & 0 & 118,979 (39.7\%) \\
    & 1 & 101,198 (33.8\%) \\
    & 2 & 35,712 (11.9\%) \\
    & 3-5 & 29,866 (10.0\%) \\
    & 6+ & 13,957 (4.7\%) \\
    \midrule
    Length of first hospital stay & \textless{}1 day & 47,804 (15.9\%) \\
    & 1-3 days & 57,392 (19.1\%) \\
    & 3-7 days & 48,229 (16.1\%) \\
    & 1-2 weeks & 18,594 (6.2\%) \\
    & \textgreater{}2 weeks & 8,647 (2.9\%) \\
    \midrule
    In-hospital mortality (any stay) & Died in hospital & 8,493 (2.8\%) \\
    & Survived hospitalization & 172,240 (57.5\%) \\
    \midrule
    Mortality (any cause, in record) & Deceased & 29,076 (9.7\%) \\
    & Alive / censored & 270,636 (90.3\%) \\
    \midrule
    \textbf{Total} & \textbf{Patients} & \textbf{299,712} \\
    \bottomrule
  \end{tabular}
\end{table}

\end{nolinenumbers}

\clearpage
\setcounter{figure}{0}
\setcounter{table}{0}
\renewcommand{\figurename}{\textbf{Supplementary Figure}}
\renewcommand{\tablename}{\textbf{Supplementary Table}}
\begin{nolinenumbers}
\newpage
\Heading{Supplementary Information}

\begin{table}[ht!]
    \centering
    \caption{\textbf{Per-event-type latent surprise.} Median, upper-tail ($q_{95}$), and mean KL$(q\|p)$ per event describing \ours's latent innovation, i.e. the surprise of the trained model on the test set. By event type, sorted by median descending. A larger value means the event type is less predictable from the patient history the model has seen up to this point. Routine measurements sit near zero, while imaging and waveform events with their rather unpredictable contents, as well as terminal events, carry the most surprise.}
    \label{tab:latent_kl_by_event_type}
    \begin{tabular}{lrrr}
        \toprule
        \textbf{Event type} & \textbf{Median KL} & \textbf{$q_{95}$ KL} & \textbf{Mean KL} \\
        \midrule
        ECG & 38 & 49.9 & 38.7 \\
        X-ray & 37 & 57.3 & 37.2 \\
        ECHO & 22.5 & 41.9 & 24 \\
        Note Radiologyreport & 2.86 & 6.45 & 3 \\
        Death & 0.961 & 21.5 & 4.24 \\
        Enter Hospitalization & 0.558 & 6.39 & 1.39 \\
        Note Dischargesummary & 0.452 & 5.96 & 1.33 \\
        Enter ICU & 0.16 & 6.65 & 1.1 \\
        Leave ICU & 0.105 & 7.95 & 1.06 \\
        Microbiology Test & 0.0605 & 3.4 & 0.75 \\
        Service & 0.0571 & 4.18 & 0.644 \\
        Leave Transfer & 0.0553 & 5.63 & 1.12 \\
        Leave Hospitalization & 0.0545 & 2.66 & 0.504 \\
        Procedure End & 0.0506 & 6.81 & 1.34 \\
        Leave ED & 0.0436 & 0.394 & 0.179 \\
        Body Output & 0.0401 & 0.235 & 0.0667 \\
        Transfer & 0.04 & 2.5 & 0.374 \\
        Drug Prescription & 0.0381 & 0.574 & 0.135 \\
        Drug Start & 0.0365 & 0.964 & 0.198 \\
        Body Input & 0.0355 & 0.0588 & 0.0384 \\
        Complaint & 0.0347 & 0.548 & 0.097 \\
        Drug Stop & 0.031 & 0.932 & 0.18 \\
        Enter ED & 0.0309 & 0.0376 & 0.0316 \\
        Drug Previously & 0.0306 & 0.51 & 0.0946 \\
        Procedure & 0.0305 & 0.0497 & 0.038 \\
        Diagnosis & 0.0286 & 0.0757 & 0.0344 \\
        Other Event & 0.0276 & 0.0457 & 0.0302 \\
        Triage Acuity & 0.0258 & 0.0339 & 0.037 \\
        Vitals & 0.0256 & 0.0526 & 0.0584 \\
        Lab Test & 0.0242 & 0.0593 & 0.0533 \\
        Body Input End & 0.0222 & 0.0544 & 0.0281 \\
        \bottomrule
    \end{tabular}
\end{table}

\begin{table}[ht!]
    \centering
    \caption{\textbf{Forecasting.} The same forecasts as Extended Data Table~\ref{tab:forecasting-headline}, evaluated at a single horizon $H{=}24$ h, i.e. scoring events that occur within 24 h after the prompt cutoff. Patients are split into quartiles over length of stay. Each cell is the macro-averaged event-type AUROC ($\uparrow$ better) or Brier score ($\downarrow$ better) for that subgroup. Computed over $N{=}13{,}551$ patients. Best model per column is in \textbf{bold}, and parentheses give 95\% patient-bootstrap CIs ($B{=}1{,}000$ resamples).}
    \label{tab:forecasting-los-subgroup}
    \setlength{\tabcolsep}{4pt}
    \resizebox{\textwidth}{!}{
        \begin{tabular}{llrrrr}
            \toprule
            {\bf Predictor} & {\bf Metric} & \bf Q1 ($\le 48$ h) & \bf Q2 (48-91 h) & \bf Q3 (91-164 h) & \bf Q4 ($> 164$ h) \\
            \midrule
            Persistence & AUROC & 0.597 (0.588, 0.609) & 0.647 (0.636, 0.656) & 0.639 (0.631, 0.647) & 0.610 (0.605, 0.631) \\
            Persistence & Brier & 0.287 (0.283, 0.290) & 0.293 (0.288, 0.297) & 0.317 (0.312, 0.322) & 0.342 (0.338, 0.348) \\
            \ours (free) & AUROC & 0.828 (0.807, 0.847) & 0.823 (0.809, 0.836) & 0.833 (0.822, 0.842) & 0.835 (0.821, 0.844) \\
            \ours (free) & Brier & \textbf{0.084} (0.082, 0.086) & 0.099 (0.096, 0.101) & 0.103 (0.100, 0.106) & 0.110 (0.107, 0.114) \\
            \ours (time-controlled) & AUROC & \textbf{0.868} (0.851, 0.882) & \textbf{0.872} (0.859, 0.885) & \textbf{0.864} (0.855, 0.874) & \textbf{0.865} (0.853, 0.873) \\
            \ours (time-controlled) & Brier & 0.096 (0.093, 0.098) & \textbf{0.085} (0.082, 0.087) & \textbf{0.095} (0.093, 0.098) & \textbf{0.106} (0.104, 0.109) \\
            \bottomrule
        \end{tabular}
    }
\end{table}

\bigskip

\begin{table}[ht!]
    \centering
    \caption{\textbf{Forecasting.} The 10 best- and 10 worst-predicted event types in terms of AUROC at the $H{=}24$ h horizon over $N{=}13{,}551$ patients. Prev. is the cohort prevalence. AUROC for free and time-controlled (TC) forecasting. Count MAE (free) is the mean absolute error between predicted and actual per-patient event counts in the window. Best rollout variant is in \textbf{bold}.}
    \label{tab:forecasting-perclass}
    \setlength{\tabcolsep}{4pt}
    \footnotesize
    \begin{tabular}{lrrrr}
        \toprule
        {\bf Class} & {\bf Prev.} & {\bf AUROC (free)} & {\bf AUROC (TC)} & {\bf Count MAE (free)} \\
        \midrule
        \multicolumn{5}{l}{\textit{Top-10 by free-mode AUROC}} \\
        \addlinespace[3pt]
        Leave ED & 1.2\% & \textbf{0.982} & 0.914 & 0.008 \\
        Leave ICU & 1.8\% & 0.967 & \textbf{0.982} & 0.045 \\
        Body Output & 20.4\% & 0.964 & \textbf{0.974} & 2.305 \\
        Other Event & 25.9\% & 0.953 & \textbf{0.979} & 3.245 \\
        Complaint & 7.2\% & 0.941 & \textbf{0.960} & 1.049 \\
        Body Input End & 19.1\% & 0.930 & \textbf{0.960} & 1.263 \\
        Body Input & 19.6\% & 0.930 & \textbf{0.961} & 1.474 \\
        Vitals & 31.8\% & 0.930 & \textbf{0.954} & 7.949 \\
        Procedure End & 5.9\% & 0.925 & \textbf{0.938} & 0.401 \\
        X-ray & 1.9\% & 0.901 & \textbf{0.902} & 0.139 \\
        \midrule
        \multicolumn{5}{l}{\textit{Bottom-10 by free-mode AUROC}} \\
        \addlinespace[3pt]
        Note Radiologyreport & 11.7\% & 0.687 & \textbf{0.735} & 0.361 \\
        Service & 1.9\% & 0.696 & \textbf{0.806} & 0.058 \\
        Transfer & 9.5\% & 0.696 & \textbf{0.777} & 0.175 \\
        Diagnosis & 31.0\% & 0.713 & \textbf{0.731} & 5.392 \\
        Note Dischargesummary & 12.1\% & 0.723 & \textbf{0.877} & 0.367 \\
        Triage Acuity & 0.1\% & \textbf{0.732} & 0.689 & 0.010 \\
        Enter ED & 0.1\% & 0.732 & \textbf{0.757} & 0.012 \\
        Leave Transfer & 24.9\% & \textbf{0.760} & 0.734 & 0.464 \\
        Lab Test & 63.6\% & 0.773 & \textbf{0.810} & 13.171 \\
        ECG & 6.2\% & 0.778 & \textbf{0.808} & 0.240 \\
        \bottomrule
    \end{tabular}
\end{table}

\begin{table}[ht!]
    \centering
    \caption{{\bf Counterfactual simulation.} Calibration of 30-day in-hospital mortality prediction under the factual intervention against the observed outcome ($n{=}1{,}085$ uncensored patients), by decile of predicted risk. The final column gives each decile's contribution to the ECE. \ours significantly tends toward overpredicting mortality for this subcohort.}
    \label{tab:counterfactual-calibration}
    \setlength{\tabcolsep}{4pt}
    \begin{tabular}{rrrrrr}
        \toprule
        {\bf Decile} & {\bf $n$} & {\bf Mean predicted} & {\bf Observed risk} & {\bf $|$gap$|$} & {\bf Bin-ECE contribution} \\
        \midrule
        1 & 108 & 3.04\% & 3.70\% & 0.67 pp & 0.066 pp \\
        2 & 109 & 13.69\% & 10.09\% & 3.60 pp & 0.361 pp \\
        3 & 108 & 25.56\% & 20.37\% & 5.19 pp & 0.516 pp \\
        4 & 109 & 35.38\% & 15.60\% & 19.78 pp & 1.987 pp \\
        5 & 108 & 43.93\% & 25.00\% & 18.93 pp & 1.884 pp \\
        6 & 109 & 51.96\% & 15.60\% & 36.37 pp & 3.653 pp \\
        7 & 108 & 60.59\% & 25.93\% & 34.67 pp & 3.451 pp \\
        8 & 109 & 70.72\% & 22.02\% & 48.70 pp & 4.892 pp \\
        9 & 108 & 84.30\% & 25.93\% & 58.37 pp & 5.810 pp \\
        10 & 109 & 96.26\% & 45.87\% & 50.39 pp & 5.062 pp \\
        \midrule
        \textbf{Total} & 1,085 & - & - & - & \textbf{27.683 pp (ECE)} \\
        \bottomrule
    \end{tabular}
\end{table}

\begin{table}[ht!]
    \centering
    \caption{{\bf Counterfactual simulation.} Fraction of Monte Carlo simulations that reach a determined outcome within the 30-day window. Per arm (intervention) with the associated invalid-run ratio.}
    \label{tab:counterfactual-coverage}
    \setlength{\tabcolsep}{4pt}
    \begin{tabular}{lrrr}
        \toprule
        {\bf Arm} & {\bf Mean coverage} & {\bf Invalid-run ratio} & {\bf 95\% CI on coverage} \\
        \midrule
        Matched & 68.57\% & 31.43\% & (66.89\%, 70.22\%) \\
        Saline & 68.48\% & 31.52\% & (66.84\%, 70.21\%) \\
        Ringer's & 71.78\% & 28.22\% & (70.21\%, 73.34\%) \\
        \bottomrule
    \end{tabular}
\end{table}

\begin{table}[ht!]
    \centering
    \caption{{\bf Counterfactual simulation.} Simulated effect of saline versus Ringer's on MAKE-30, the primary endpoint of the SMART trial (i.e. occurrence of death, new renal-replacement therapy, or a final serum creatinine $\geq$2$\times$ baseline within 30 days). Rows as in Extended Data Table~\ref{tab:counterfactual-headline}. Observed mortality counts any death within 30 d, following the MAKE-30 definition. The renal component is evaluated only where the simulation emits a creatinine and a baseline exists. Reference: all SMART patients \cite{counterfactual-medical-study} and sepsis subgroup \cite{counterfactual-medical-study-sepsis} with MAKE-30 at $40.1$ vs. $35.4$\% for saline vs. balanced crystalloid, adjusted OR $0.78$ ($95$\% CI $0.63$-$0.97$), interval derived from the published arm rates (unadjusted).}
    \label{tab:cf-make30}
    \setlength{\tabcolsep}{4pt}
    \begin{tabular}{lcccc}
        \toprule
         & {\bf MAKE-30} & \multicolumn{3}{c}{\bf Components (\%)} \\
        \cmidrule(lr){2-2} \cmidrule(lr){3-5}
        {\bf Scenario} & {\bf composite (\%)} & {\it death} & {\it new RRT} & {\it renal dysf.} \\
        \midrule
        Observed & $27.0$ ($24.4$, $29.6$) & $23.4$ ($20.8$, $25.7$) & $5.6$ ($4.3$, $7.0$) & $3.7$ ($2.6$, $4.9$) \\
        Matched {\footnotesize(calibration)} & $73.8$ ($71.9$, $75.4$) & $73.7$ ($71.9$, $75.4$) & $0.2$ ($0.1$, $0.3$) & $0.2$ ($0.1$, $0.3$) \\
        Saline & $74.7$ ($72.8$, $76.6$) & $74.6$ ($72.9$, $76.3$) & $0.2$ ($0.1$, $0.3$) & $0.2$ ($0.1$, $0.3$) \\
        Ringer's & $65.8$ ($63.9$, $67.8$) & $65.8$ ($63.9$, $67.9$) & $0.1$ ($0.0$, $0.1$) & $0.0$ ($0.0$, $0.1$) \\
        \midrule
        {\bf Effect} {\footnotesize (Saline $-$ Ringer's)} & $+8.92$ pp ($7.85$, $9.93$) & $+8.86$ pp ($7.83$, $9.93$) & $+.13$ pp ($.04$, $.23$) & $+.13$ pp ($.05$, $.22$) \\
        SMART, sepsis \cite{counterfactual-medical-study-sepsis} & $+4.7$ pp ($+0.1$, $+9.4$) & - & - & - \\
        SMART, all \cite{counterfactual-medical-study} & $+1.1$ pp ($+0.2$, $+1.9$) & - & - & - \\
        \bottomrule
    \end{tabular}
\end{table}

\begin{table}[ht!]
  \centering
  \setlength{\tabcolsep}{14pt}\renewcommand{\arraystretch}{1.15}
  \caption{\textbf{Linear probing.} AUROC of $\ell_2$-regularized logistic probes that decode clinical outcomes from the frozen patient state embeddings at fixed points along the last stay (mean $\pm$ s.d., 5-fold $\times$ 3-seed stratified cross-validation, one probe per position). The probe at discharge is a clean retrieval probe, after the outcome is already known to the model and thus potentially to the patient embedding used for this probe. For the other timepoints, the kind and degree of information available to the model at each timestep when producing the embeddings vary between patients and stays. Thus, the embeddings up to a certain timepoint do not include explicit or implicit information about the label and their probes are therefore \emph{predictive} rather than \emph{retrieving}. However, this point is individual for each patient and stay, so we consider all probes during the stay as read-outs that mix prediction and retrieval. For each of the temporal positions within the stay, we only score stays long enough to reach them. Thus, the number of samples naturally shrinks over rows (total number of samples: hospital $N=27{,}185$, ICU $N=7{,}479$).}
  \label{tab:lp-outcomes}
  \begin{tabular}{lcccc}
    \toprule
     & \multicolumn{2}{c}{\textbf{Length of stay $\geq$ 72 h}} & \multicolumn{2}{c}{\textbf{In-hospital mortality}} \\
    \cmidrule(lr){2-3} \cmidrule(lr){4-5}
    \textbf{Position} & \textbf{Hospital} & \textbf{ICU} & \textbf{$\leq$ 72 h} & \textbf{Whole stay} \\
    \midrule
    \multicolumn{5}{l}{\textit{Read-out, prediction/retrieval mixed (State during the stay)}} \\
    Pre-admission & $0.836 \pm 0.003$ & $0.688 \pm 0.012$ & $0.907 \pm 0.022$ & $0.900 \pm 0.007$ \\
    At admission & $0.860 \pm 0.005$ & $0.712 \pm 0.011$ & $0.919 \pm 0.020$ & $0.913 \pm 0.007$ \\
    $+$12 h & $0.867 \pm 0.005$ & $0.754 \pm 0.010$ & $0.914 \pm 0.016$ & $\mathbf{0.914} \pm 0.007$ \\
    $+$24 h & $0.861 \pm 0.005$ & $\mathbf{0.784} \pm 0.012$ & $0.919 \pm 0.014$ & $0.896 \pm 0.010$ \\
    $+$36 h & $0.888 \pm 0.004$ & $0.764 \pm 0.014$ & $0.926 \pm 0.025$ & $0.897 \pm 0.011$ \\
    $+$48 h & $\mathbf{0.925} \pm 0.004$ & $0.770 \pm 0.020$ & $\mathbf{0.932} \pm 0.025$ & $0.881 \pm 0.012$ \\
    \addlinespace
    \multicolumn{5}{l}{\textit{Clean retrieval (State at discharge)}} \\
    At discharge & $0.977 \pm 0.002$ & $0.871 \pm 0.009$ & $0.993 \pm 0.002$ & $0.998 \pm 0.003$ \\
    \bottomrule
  \end{tabular}
\end{table}

\begin{table}[ht!]
  \centering
  \setlength{\tabcolsep}{10pt}\renewcommand{\arraystretch}{1.15}
  \caption{\textbf{Linear probing.} Read-out and retrieval of the ICD chapters of a stay from the patient state embedding at the beginning and end of the stay. AUROC (mean $\pm$ s.d. over 5-fold $\times$ 3-seed stratified cross-validation) of a logistic probe on the states at admission, at $+48$ h, and at discharge (clean retrieval). The $+48$ h column is restricted to the subcohort of stays that reach it. Rows sorted by discharge AUROC descending.}
  \label{tab:lp-icd}
  \begin{tabular}{l r c c c}
    \toprule
    & & \multicolumn{2}{c}{\textbf{Read-out}} & \textbf{Retrieval} \\
    \cmidrule(lr){3-4} \cmidrule(lr){5-5}
    \textbf{Target} & \textbf{Prev.\,(\%)} & \textbf{Admission} & \textbf{$+$48 h} & \textbf{Discharge} \\
    \midrule
    Circulatory & 58.2 & $0.913 \pm 0.004$ & $0.907 \pm 0.004$ & $0.937 \pm 0.003$ \\
    Neoplasms & 14.1 & $0.828 \pm 0.009$ & $0.768 \pm 0.010$ & $0.887 \pm 0.007$ \\
    Endocrine / metabolic & 55.8 & $0.837 \pm 0.005$ & $0.815 \pm 0.007$ & $0.878 \pm 0.005$ \\
    Injury / poisoning & 27.4 & $0.791 \pm 0.004$ & $0.729 \pm 0.010$ & $0.868 \pm 0.007$ \\
    Genitourinary & 29.9 & $0.817 \pm 0.005$ & $0.793 \pm 0.005$ & $0.865 \pm 0.004$ \\
    Factors infl. health (Z/V) & 65.4 & $0.820 \pm 0.005$ & $0.783 \pm 0.006$ & $0.864 \pm 0.004$ \\
    Infectious & 17.1 & $0.803 \pm 0.006$ & $0.778 \pm 0.009$ & $0.857 \pm 0.007$ \\
    Mental / behavioral & 40.9 & $0.745 \pm 0.007$ & $0.706 \pm 0.005$ & $0.850 \pm 0.006$ \\
    Respiratory & 27.3 & $0.769 \pm 0.005$ & $0.766 \pm 0.007$ & $0.838 \pm 0.007$ \\
    Digestive / hepatic & 37.3 & $0.777 \pm 0.005$ & $0.735 \pm 0.008$ & $0.831 \pm 0.005$ \\
    Symptoms / signs & 52.7 & $0.783 \pm 0.006$ & $0.762 \pm 0.008$ & $0.814 \pm 0.006$ \\
    Skin & 9.7 & $0.755 \pm 0.011$ & $0.709 \pm 0.012$ & $0.800 \pm 0.007$ \\
    Nervous system & 28.9 & $0.738 \pm 0.006$ & $0.704 \pm 0.010$ & $0.792 \pm 0.004$ \\
    Musculoskeletal & 26.7 & $0.726 \pm 0.008$ & $0.691 \pm 0.009$ & $0.764 \pm 0.006$ \\
    Eye / ear (sensory) & 7.5 & $0.696 \pm 0.010$ & $0.643 \pm 0.016$ & $0.747 \pm 0.010$ \\
    \bottomrule
  \end{tabular}
\end{table}

\begin{table}[ht!]
  \centering
  \setlength{\tabcolsep}{10pt}\renewcommand{\arraystretch}{1.15}
  \caption{\textbf{Linear probing.} Read-out and retrieval of Quan-Elixhauser comorbidities of a stay from the patient state embedding at the beginning and end of the stay. AUROC (mean $\pm$ s.d. over 5-fold $\times$ 3-seed stratified cross-validation) of a logistic probe on the states at admission, at $+48$ h, and at discharge (clean retrieval). The $+48$ h column is restricted to the subcohort of stays that reach it. Rows sorted by discharge AUROC descending.}
  \label{tab:lp-elix}
  \begin{tabular}{l r c c c}
    \toprule
    & & \multicolumn{2}{c}{\textbf{Read-out}} & \textbf{Retrieval} \\
    \cmidrule(lr){3-4} \cmidrule(lr){5-5}
    \textbf{Target} & \textbf{Prev.\,(\%)} & \textbf{Admission} & \textbf{$+$48 h} & \textbf{Discharge} \\
    \midrule
    Metastatic cancer & 4.4 & $0.880 \pm 0.010$ & $0.840 \pm 0.012$ & $0.945 \pm 0.008$ \\
    Heart failure & 11.9 & $0.907 \pm 0.004$ & $0.881 \pm 0.006$ & $0.934 \pm 0.003$ \\
    Psychoses & 2.7 & $0.812 \pm 0.020$ & $0.764 \pm 0.036$ & $0.914 \pm 0.012$ \\
    Alcohol abuse & 8.4 & $0.861 \pm 0.007$ & $0.766 \pm 0.013$ & $0.910 \pm 0.007$ \\
    Renal failure & 11.0 & $0.879 \pm 0.006$ & $0.838 \pm 0.009$ & $0.910 \pm 0.005$ \\
    Diabetes compl. & 6.9 & $0.861 \pm 0.008$ & $0.827 \pm 0.010$ & $0.905 \pm 0.008$ \\
    Weight loss & 4.4 & $0.811 \pm 0.012$ & $0.776 \pm 0.015$ & $0.887 \pm 0.010$ \\
    Fluid / electrolyte & 17.5 & $0.827 \pm 0.006$ & $0.812 \pm 0.008$ & $0.885 \pm 0.004$ \\
    Valvular disease & 7.1 & $0.847 \pm 0.007$ & $0.821 \pm 0.012$ & $0.883 \pm 0.006$ \\
    Cardiac arrhythmias & 20.2 & $0.838 \pm 0.007$ & $0.821 \pm 0.005$ & $0.882 \pm 0.006$ \\
    Hypertension & 44.6 & $0.858 \pm 0.004$ & $0.825 \pm 0.007$ & $0.882 \pm 0.004$ \\
    Paralysis & 2.1 & $0.808 \pm 0.013$ & $0.777 \pm 0.030$ & $0.879 \pm 0.021$ \\
    Drug abuse & 5.2 & $0.815 \pm 0.011$ & $0.763 \pm 0.019$ & $0.877 \pm 0.009$ \\
    Pulmonary circulation & 4.0 & $0.801 \pm 0.011$ & $0.751 \pm 0.017$ & $0.862 \pm 0.009$ \\
    Liver disease & 6.1 & $0.809 \pm 0.010$ & $0.782 \pm 0.014$ & $0.861 \pm 0.009$ \\
    AIDS / HIV & 0.5 & $0.787 \pm 0.037$ & $0.717 \pm 0.056$ & $0.859 \pm 0.036$ \\
    Coagulopathy & 7.3 & $0.791 \pm 0.009$ & $0.762 \pm 0.011$ & $0.855 \pm 0.009$ \\
    Obesity & 8.3 & $0.808 \pm 0.008$ & $0.755 \pm 0.012$ & $0.849 \pm 0.008$ \\
    Depression & 16.2 & $0.711 \pm 0.009$ & $0.660 \pm 0.014$ & $0.846 \pm 0.006$ \\
    Peripheral vascular & 5.7 & $0.801 \pm 0.010$ & $0.733 \pm 0.010$ & $0.830 \pm 0.010$ \\
    Lymphoma & 1.3 & $0.778 \pm 0.030$ & $0.735 \pm 0.037$ & $0.828 \pm 0.029$ \\
    Solid tumor (non-met.) & 3.9 & $0.797 \pm 0.016$ & $0.700 \pm 0.028$ & $0.822 \pm 0.014$ \\
    Other neurological & 8.8 & $0.784 \pm 0.011$ & $0.746 \pm 0.011$ & $0.821 \pm 0.009$ \\
    Diabetes uncompl. & 11.9 & $0.755 \pm 0.008$ & $0.721 \pm 0.008$ & $0.819 \pm 0.006$ \\
    Deficiency anemia & 7.3 & $0.731 \pm 0.011$ & $0.678 \pm 0.015$ & $0.810 \pm 0.010$ \\
    Blood-loss anemia & 0.6 & $0.725 \pm 0.029$ & $0.606 \pm 0.036$ & $0.801 \pm 0.037$ \\
    Hypothyroidism & 10.4 & $0.723 \pm 0.009$ & $0.654 \pm 0.014$ & $0.778 \pm 0.011$ \\
    Chronic pulmonary & 15.9 & $0.709 \pm 0.008$ & $0.673 \pm 0.014$ & $0.777 \pm 0.008$ \\
    Rheumatoid / collagen & 2.7 & $0.680 \pm 0.013$ & $0.620 \pm 0.021$ & $0.717 \pm 0.028$ \\
    \bottomrule
  \end{tabular}
\end{table}

\clearpage
{
    \setlength{\tabcolsep}{6pt}
    \renewcommand{\arraystretch}{1.1}
    \begin{longtable}[ht!]{lcccc|cccc}
        \caption{\textbf{Linear probing.} Result of the retrieval at the end of the stay split by subpopulation. For every target, the last patient state embedding in the stay is probed separately within subgroups of age and length of stay. This investigates whether \ours carries clinical information well over time and across the cohort. Each value is the AUROC of a single retrieval probe per target. A dash marks subgroups that are not evaluable, and the best subgroup per row is in bold. Rows are sorted by mean AUROC.}
        \label{tab:lp-cohort}\\
        \toprule
         & \multicolumn{4}{c}{\textbf{Age (yr)}} & \multicolumn{4}{c}{\textbf{LOS quartile}} \\
        \cmidrule(lr){2-5} \cmidrule(lr){6-9}
        \textbf{Target} & \textbf{$<$50} & \textbf{50-65} & \textbf{65-80} & \textbf{80+} & \shortstack{\textbf{Q1}\\[3pt]0.0-1.0 d} & \shortstack{\textbf{Q2}\\[3pt]1.0-2.6 d} & \shortstack{\textbf{Q3}\\[3pt]2.6-5.1 d} & \shortstack{\textbf{Q4}\\[3pt]5.1-220.0 d} \\
        \midrule
        \endfirsthead
        \multicolumn{9}{c}{\tablename\ \thetable{} (continued)} \\
        \toprule
         & \multicolumn{4}{c}{\textbf{Age (yr)}} & \multicolumn{4}{c}{\textbf{LOS quartile}} \\
        \cmidrule(lr){2-5} \cmidrule(lr){6-9}
        \textbf{Target} & \textbf{$<$50} & \textbf{50-65} & \textbf{65-80} & \textbf{80+} & \shortstack{\textbf{Q1}\\[3pt]0.0-1.0 d} & \shortstack{\textbf{Q2}\\[3pt]1.0-2.6 d} & \shortstack{\textbf{Q3}\\[3pt]2.6-5.1 d} & \shortstack{\textbf{Q4}\\[3pt]5.1-220.0 d} \\
        \midrule
        \endhead
        \midrule \multicolumn{9}{r}{\footnotesize continued on next page} \\
        \endfoot
        \bottomrule
        \endlastfoot
        \addlinespace[10pt]
        \multicolumn{9}{l}{\textit{Clinical outcomes}} \\[4pt]
        In-hospital mortality & \textbf{0.999} & 0.999 & 0.998 & 0.999 & 0.996 & 0.999 & 0.997 & 0.999 \\
        Hospital LOS $\geq$ 72 h & 0.978 & \textbf{0.981} & 0.977 & 0.963 & - & - & 0.825 & - \\
        ICU LOS $\geq$ 72 h & 0.898 & 0.866 & 0.875 & 0.867 & \textbf{0.916} & 0.871 & 0.859 & 0.859 \\
        \addlinespace[10pt]
        \multicolumn{9}{l}{\textit{ICD chapters}} \\[4pt]
        Circulatory & 0.895 & 0.873 & 0.884 & 0.891 & 0.925 & 0.932 & \textbf{0.940} & 0.924 \\
        Neoplasms & 0.883 & \textbf{0.907} & 0.881 & 0.812 & 0.890 & 0.888 & 0.882 & 0.861 \\
        Injury / poisoning & \textbf{0.912} & 0.864 & 0.835 & 0.804 & 0.887 & 0.888 & 0.863 & 0.801 \\
        Genitourinary & 0.846 & 0.846 & 0.828 & 0.805 & 0.841 & \textbf{0.862} & 0.848 & 0.839 \\
        Factors infl. health (Z/V) & \textbf{0.867} & 0.830 & 0.834 & 0.806 & 0.830 & 0.850 & 0.862 & 0.833 \\
        Mental / behavioral & \textbf{0.911} & 0.845 & 0.809 & 0.749 & 0.902 & 0.857 & 0.839 & 0.786 \\
        Infectious & 0.848 & 0.856 & \textbf{0.868} & 0.843 & 0.788 & 0.831 & 0.836 & 0.825 \\
        Endocrine / metabolic & 0.861 & 0.835 & 0.798 & 0.744 & \textbf{0.876} & 0.863 & 0.865 & 0.842 \\
        Respiratory & 0.834 & \textbf{0.839} & 0.818 & 0.810 & 0.813 & 0.828 & 0.828 & 0.810 \\
        Digestive / hepatic & \textbf{0.879} & 0.820 & 0.780 & 0.743 & 0.846 & 0.836 & 0.820 & 0.776 \\
        Symptoms / signs & \textbf{0.836} & 0.813 & 0.782 & 0.761 & 0.822 & 0.817 & 0.813 & 0.784 \\
        Skin & \textbf{0.849} & 0.810 & 0.777 & 0.709 & 0.833 & 0.811 & 0.805 & 0.720 \\
        Nervous system & \textbf{0.832} & 0.777 & 0.748 & 0.727 & 0.804 & 0.768 & 0.776 & 0.744 \\
        Musculoskeletal & \textbf{0.803} & 0.749 & 0.697 & 0.657 & 0.768 & 0.778 & 0.790 & 0.688 \\
        Eye / ear (sensory) & \textbf{0.780} & 0.719 & 0.667 & 0.636 & 0.769 & 0.766 & 0.749 & 0.671 \\
        \addlinespace[10pt]
        \multicolumn{9}{l}{\textit{Elixhauser comorbidities}} \\[4pt]
        Metastatic cancer & 0.965 & 0.949 & 0.940 & 0.885 & \textbf{0.966} & 0.944 & 0.952 & 0.917 \\
        Heart failure & \textbf{0.955} & 0.924 & 0.894 & 0.873 & 0.931 & 0.937 & 0.935 & 0.903 \\
        Diabetes compl. & \textbf{0.957} & 0.902 & 0.873 & 0.849 & 0.897 & 0.919 & 0.930 & 0.868 \\
        Psychoses & \textbf{0.958} & 0.899 & 0.871 & 0.778 & 0.950 & 0.915 & 0.909 & 0.888 \\
        Renal failure & \textbf{0.947} & 0.905 & 0.866 & 0.802 & 0.920 & 0.923 & 0.915 & 0.863 \\
        Alcohol abuse & 0.948 & 0.882 & 0.822 & 0.750 & \textbf{0.958} & 0.902 & 0.916 & 0.832 \\
        Weight loss & \textbf{0.930} & 0.899 & 0.870 & 0.804 & 0.902 & 0.879 & 0.881 & 0.813 \\
        Paralysis & \textbf{0.890} & 0.881 & 0.864 & 0.868 & 0.849 & 0.847 & 0.885 & 0.845 \\
        AIDS / HIV & 0.923 & 0.772 & 0.875 & - & 0.811 & \textbf{0.934} & 0.914 & 0.820 \\
        Fluid / electrolyte & \textbf{0.906} & 0.894 & 0.860 & 0.808 & 0.847 & 0.888 & 0.866 & 0.830 \\
        Valvular disease & \textbf{0.922} & 0.857 & 0.833 & 0.786 & 0.862 & 0.874 & 0.879 & 0.857 \\
        Cardiac arrhythmias & 0.833 & 0.853 & 0.845 & 0.814 & 0.867 & \textbf{0.885} & 0.883 & 0.846 \\
        Pulmonary circulation & \textbf{0.940} & 0.848 & 0.799 & 0.792 & 0.903 & 0.868 & 0.849 & 0.783 \\
        Liver disease & \textbf{0.905} & 0.861 & 0.810 & 0.773 & 0.843 & 0.853 & 0.858 & 0.844 \\
        Obesity & \textbf{0.896} & 0.852 & 0.791 & 0.806 & 0.870 & 0.879 & 0.856 & 0.793 \\
        Depression & 0.901 & 0.839 & 0.813 & 0.756 & \textbf{0.902} & 0.867 & 0.841 & 0.778 \\
        Drug abuse & 0.874 & 0.822 & 0.798 & 0.679 & \textbf{0.908} & 0.904 & 0.871 & 0.834 \\
        Hypertension & 0.900 & 0.809 & 0.757 & 0.696 & \textbf{0.904} & 0.896 & 0.886 & 0.811 \\
        Coagulopathy & \textbf{0.897} & 0.870 & 0.827 & 0.773 & 0.842 & 0.840 & 0.812 & 0.793 \\
        Lymphoma & \textbf{0.868} & 0.843 & 0.815 & 0.734 & 0.847 & 0.815 & 0.829 & 0.805 \\
        Blood-loss anemia & 0.900 & 0.853 & 0.728 & 0.716 & \textbf{0.920} & 0.886 & 0.805 & 0.679 \\
        Solid tumor (non-met.) & \textbf{0.923} & 0.823 & 0.772 & 0.684 & 0.861 & 0.853 & 0.815 & 0.752 \\
        Peripheral vascular & \textbf{0.885} & 0.823 & 0.763 & 0.700 & 0.883 & 0.850 & 0.812 & 0.758 \\
        Diabetes uncompl. & \textbf{0.885} & 0.792 & 0.753 & 0.742 & 0.856 & 0.858 & 0.828 & 0.741 \\
        Other neurological & \textbf{0.863} & 0.823 & 0.803 & 0.730 & 0.800 & 0.831 & 0.828 & 0.763 \\
        Deficiency anemia & 0.858 & 0.818 & 0.771 & 0.738 & \textbf{0.859} & 0.829 & 0.794 & 0.730 \\
        Chronic pulmonary & \textbf{0.804} & 0.773 & 0.741 & 0.719 & 0.800 & 0.798 & 0.773 & 0.715 \\
        Hypothyroidism & \textbf{0.844} & 0.730 & 0.707 & 0.689 & 0.809 & 0.806 & 0.782 & 0.710 \\
        Rheumatoid / collagen & \textbf{0.822} & 0.707 & 0.643 & 0.592 & 0.737 & 0.722 & 0.743 & 0.634 \\
    \end{longtable}
}

\begin{figure}[ht!]
    \centering
    \includegraphics[width=\linewidth]{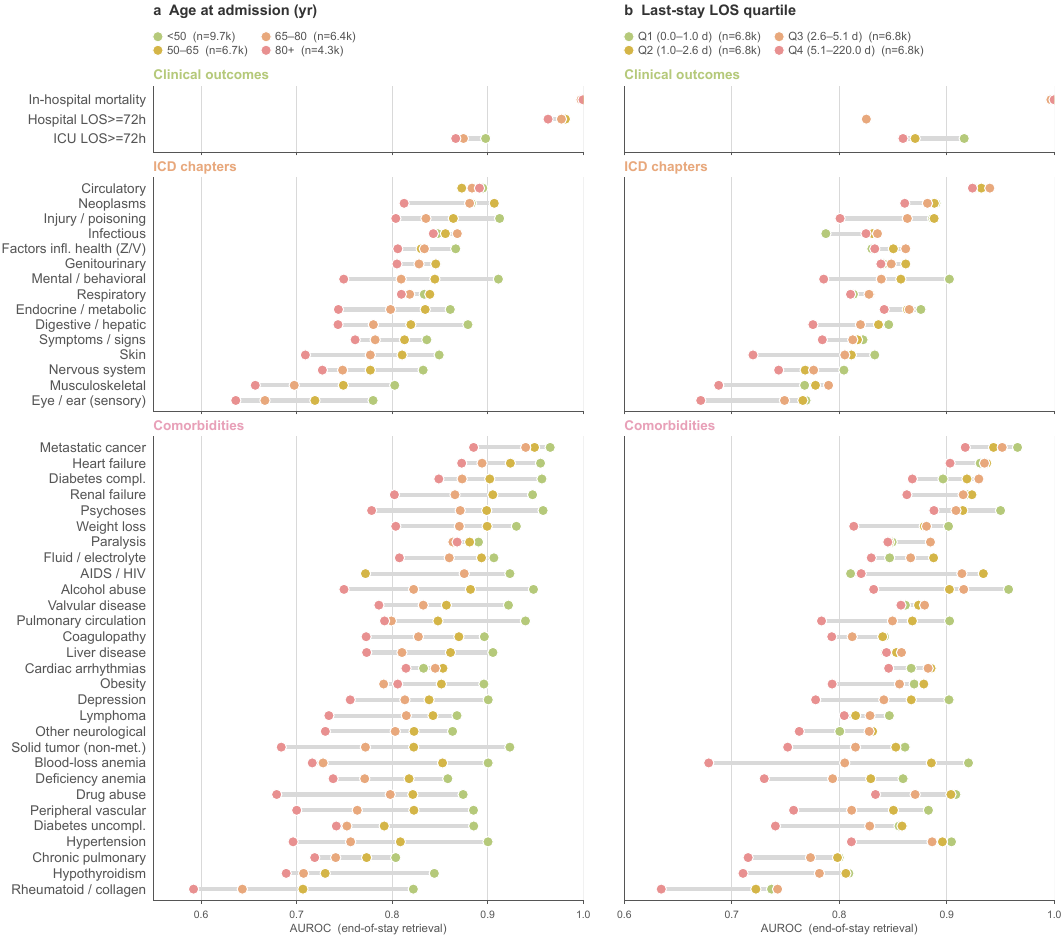}
    \caption{\textbf{Linear probing of the final patient state for retrieval.} Results split by age at admission and by length of hospital stay. For details on the probing setup, see Section~\textbf{Online Methods} above and the caption of Extended Data Fig.~\ref{fig:res_lp_tmp}. We observe that the probe performs better on younger patients and shorter stays across most tasks.}
     \label{fig:res_lp_cohort}
\end{figure}

\clearpage

\begin{table}[ht!]
    \centering
    \small
    \setlength{\tabcolsep}{6pt}\renewcommand{\arraystretch}{1.15}
    \caption{\textbf{NEWS2 scoring.} Recovery of the NEWS2 score from \ours's patient state embeddings on the held-out test patients ($39{,}952$ windows, $4{,}414$ patients, mean ground truth NEWS2 is 4.25). Linear regression is done using ordinary least squares with an $\ell_2$ penalty. $R^2_{\mathrm{within}}$ normalizes the error by the within-patient label variance, so $0$ means no better than that patient's own mean score. $\rho$ is Spearman over windows for the window-level results and it is the mean within-patient Pearson for the patient macro results. $\kappa_w$ is quadratic-weighted over the three clinical bands. Square brackets are 95\% percentile intervals from resampling patients with replacement.}
    \label{tab:news2-headline}
    \begin{tabular}{lccccc}
        \toprule
        & \multicolumn{5}{c}{\textbf{Predictor}} \\
        \cmidrule(lr){2-6}
        & \makecell{\ours MLP\\512-256} & \makecell{constant\\(train mean)} & \makecell{constant\\(train median)} & \makecell{linear\\regression ($\ell_2$),\\age only} & \makecell{linear\\regression ($\ell_2$),\\768-dimensional\\state} \\
        \midrule
        \multicolumn{6}{l}{\textit{Window level}} \\
        \textbf{MAE} & \makecell{1.336\\{\scriptsize[1.30, 1.36]}} & \makecell{2.958\\{\scriptsize[2.90, 3.02]}} & \makecell{2.939\\{\scriptsize[2.87, 3.00]}} & \makecell{2.841\\{\scriptsize[2.78, 2.91]}} & \makecell{1.509\\{\scriptsize[1.48, 1.54]}} \\
        \textbf{RMSE} & \makecell{1.784\\{\scriptsize[1.75, 1.82]}} & \makecell{3.436\\{\scriptsize[3.37, 3.50]}} & \makecell{3.443\\{\scriptsize[3.37, 3.52]}} & \makecell{3.355\\{\scriptsize[3.28, 3.43]}} & \makecell{1.958\\{\scriptsize[1.92, 1.99]}} \\
        $\mathbf{R^2}$ & \makecell{0.730\\{\scriptsize[0.72, 0.74]}} & \makecell{-0.001\\{\scriptsize[-0.01, -0.00]}} & \makecell{-0.005\\{\scriptsize[-0.02, -0.00]}} & \makecell{0.045\\{\scriptsize[0.02, 0.06]}} & \makecell{0.675\\{\scriptsize[0.66, 0.69]}} \\
        $\mathbf{R^2_{\mathrm{within}}}$ & \makecell{0.252\\{\scriptsize[0.22, 0.29]}} & \makecell{-1.669\\{\scriptsize[-1.85, -1.52]}} & \makecell{-1.702\\{\scriptsize[-1.88, -1.54]}} & \makecell{-1.565\\{\scriptsize[-1.75, -1.41]}} & \makecell{0.103\\{\scriptsize[0.06, 0.14]}} \\
        $\boldsymbol{\rho}$ & 0.862 & - & - & 0.220 & 0.836 \\
        \textbf{macro F1} & 0.658 & 0.230 & 0.230 & 0.315 & 0.637 \\
        $\boldsymbol{\kappa_w}$ & 0.757 & 0.000 & 0.000 & 0.145 & 0.733 \\
        \textbf{AUROC$_{\geq 7}$} & 0.905 & 0.500 & 0.500 & 0.589 & 0.891 \\
        \midrule
        \multicolumn{6}{l}{\textit{Patient macro}} \\
        \textbf{MAE} & 0.814 & 3.423 & 3.148 & 3.111 & 1.062 \\
        \textbf{RMSE} & 0.940 & 3.545 & 3.278 & 3.240 & 1.210 \\
        $\boldsymbol{\rho}$ & 0.483 & 0.000 & - & -0.066 & 0.335 \\
        \bottomrule
    \end{tabular}
\end{table}

\begin{table}[ht!]
  \centering
  \small
  \setlength{\tabcolsep}{6pt}\renewcommand{\arraystretch}{1.15}
  \caption{\textbf{NEWS2 scoring.} Top: recovery of the three clinical response bands of the NEWS2 score from the rounded, clipped prediction by the MLP. Bottom: error per true integer score.}
  \label{tab:news2-breakdown}
  \begin{tabular}{lrrrr}
    \toprule
    \textbf{Risk band} & \textbf{Windows} & \textbf{Share} & \textbf{Precision} & \textbf{Recall} \\
    \midrule
    low (0-4) & 21{,}004 & 52.6\% & 0.889 & 0.852 \\
    medium (5-6) & 7{,}740 & 19.4\% & 0.387 & 0.411 \\
    high ($\geq$7) & 11{,}208 & 28.1\% & 0.693 & 0.718 \\
    \bottomrule
  \end{tabular}
  \par\vspace{8pt}
  \begin{tabular}{rrrrr}
    \toprule
    \textbf{True score} & \textbf{Windows} & \textbf{Mean $\hat{y}$} & \textbf{MAE} & \textbf{Bias} \\
    \midrule
    0 & 7{,}657 & 0.80 & 0.81 & 0.80 \\
    1 & 4{,}394 & 1.42 & 0.79 & 0.42 \\
    2 & 3{,}414 & 2.87 & 1.48 & 0.87 \\
    3 & 3{,}019 & 3.74 & 1.55 & 0.74 \\
    4 & 2{,}520 & 4.51 & 1.45 & 0.51 \\
    5 & 4{,}107 & 5.63 & 1.46 & 0.63 \\
    6 & 3{,}633 & 6.11 & 1.32 & 0.11 \\
    7 & 3{,}261 & 6.64 & 1.24 & -0.36 \\
    8 & 2{,}955 & 7.06 & 1.42 & -0.94 \\
    9 & 2{,}116 & 7.53 & 1.75 & -1.47 \\
    10 & 1{,}298 & 8.01 & 2.12 & -1.99 \\
    11 & 793 & 8.41 & 2.65 & -2.59 \\
    12 & 443 & 9.00 & 3.07 & -3.00 \\
    13 & 182 & 9.41 & 3.60 & -3.59 \\
    14 & 95 & 9.98 & 4.02 & -4.02 \\
    15 & 46 & 10.40 & 4.60 & -4.60 \\
    16 & 11 & 10.28 & 5.72 & -5.72 \\
    17 & 6 & 11.54 & 5.46 & -5.46 \\
    18 & 2 & 13.73 & 4.27 & -4.27 \\
    \bottomrule
  \end{tabular}
\end{table}

\begin{table}[ht!]
    \centering
    \small
    \caption{\textbf{Time-to-event prediction.} Survival analysis for within-stay time fractions evaluated using time-dependent C-index (td-CI), integrated Brier score (IBS), and D-calibration (D-cal) p-values. For td-CI and IBS, the mean $\pm$ s.d.\ across the five folds is reported. D-calibration is reported as the p-value from each outer fold.}
    \label{tab:deepsurv_results}
    \begin{tabular}{lccc}
        \toprule
        Within-stay time fraction $p$ & td-CI $\uparrow$ & IBS $\downarrow$ & D-cal p-values ($< 0.05 \rightarrow$ miscalibration)  \\
        \midrule
        0.0  & 0.760 $\pm$ 0.005 & 0.023 $\pm$ 0.005 & \{0.860, 0.994, 0.998, 0.994, 0.994\} \\
        \addlinespace
        0.2  & 0.784 $\pm$ 0.018 & 0.022 $\pm$ 0.006 & \{0.920, 0.868, 0.882, 0.963, 0.594\} \\
        \addlinespace
        0.4  & 0.793 $\pm$ 0.014 & 0.021 $\pm$ 0.005 & \{0.864, 0.945, 0.974, 0.854, 0.983\} \\
        \addlinespace
        0.6  & 0.813 $\pm$ 0.014 & 0.023 $\pm$ 0.005 & \{0.301, 0.561, 0.993, 0.925, 0.943\} \\
        \addlinespace
        0.8 & 0.830 $\pm$ 0.011 & 0.023 $\pm$ 0.009 & \{0.337, 0.310, 0.958, 0.200, 0.635\} \\
        \addlinespace
        1.0 & 0.849 $\pm$ 0.013 & 0.024 $\pm$ 0.026 & \{0.603, 0.183, 0.383, 0.417, 0.054\} \\
        \bottomrule
    \end{tabular}
\end{table}
\begin{table}[ht!]
    \centering
    \caption{\textbf{Survival analysis.} Summary of the hyperparameter search space.}
    \label{tab:hyparams:tte}
    \footnotesize
    \setlength{\tabcolsep}{4pt}
    \renewcommand{\arraystretch}{1.0}
    \begin{tabular}{llr}
        \toprule
        \textbf{Model} & \textbf{Hyperparameter} & \textbf{Domain} \\
        \midrule
        \textit{DeepSurv}~\cite{katzman18} & Optimizer & \{Adam, SGD\} \\
         & Weight decay & [1e{-}6, 1e{-}2] \\
         & Momentum & [0.0, 0.9] \\
         & Learning rate & [1e{-}5, 1e{-}2] \\
         & Dropout & [0.0, 0.5] \\
         & Layers & layers = [
            [width] * depth \\ & & 
            for width in \{32, 64, 128, 256\} \\ & & 
            for depth in [1, 6]
        ] \\
         \bottomrule
    \end{tabular}
\end{table}

\begin{table}[ht!]
  \centering
  \small
  \setlength{\tabcolsep}{4pt}
  \caption{\textbf{Dataset.} Event volume statistics per modality along with the value embedding dimensionality where applicable. Pt.\,cov.\ is the fraction of cohort patients with at least one event of that modality.}
  \label{tab:dataset-modality-statistics}
  \begin{tabular}{@{}lrrrrrr@{}}
    \toprule
    \textbf{Modality} & \textbf{Total} & \textbf{Train} & \textbf{Val} & \textbf{Test} & \textbf{Dim} & \textbf{Pt.\,cov.} \\
    \midrule
    Categorical & 82M & 57M & 12M & 12M & - & 100.0\% \\
    Numeric & 272M & 190M & 41M & 41M & - & 94.9\% \\
    Imaging (CXR) & 357k & 250k & 53k & 53k & 768 & 20.6\% \\
    Text & 204M & 143M & 31M & 30M & 768 & 93.3\% \\
    ECG & 795k & 556k & 118k & 119k & 192 & 53.3\% \\
    ECHO & 281k & 198k & 39k & 43k & 512 & 1.5\% \\
    \midrule
    \textbf{Total} & \textbf{559M} & \textbf{391M} & \textbf{84M} & \textbf{83M} & - & \textbf{299k pts} \\
    \bottomrule
  \end{tabular}
\end{table}

\begin{table}[ht!]
  \centering
  \caption{\textbf{Hyperparameter configuration for pretraining \ours.} Training was conducted on 8$\times$ NVIDIA A100 80GB GPUs.}
  \label{tab:hyperparameters}
  \footnotesize
  \setlength{\tabcolsep}{4pt}
  \renewcommand{\arraystretch}{1.0}
  \begin{tabular}{@{}lr@{}}
    \toprule
    \textbf{Hyperparameter} & \textbf{Value} \\
    \midrule
    \addlinespace[2pt]
    \multicolumn{2}{@{}l}{\textit{Architecture}} \\
    \addlinespace[1pt]
    \quad Trainable parameters & 150,322,970 \\
    \quad Hidden dimension & 768 \\
    \quad Transformer layers & 12 \\
    \quad Attention heads & 12 \\
    \quad FFN expansion ratio & 8 \\
    \quad Latent bottleneck dimension & 384 \\
    \quad Context window (tokens) & 2,048 \\
    \quad Attention mask & Causal (autoregressive) \\
    \quad Position encoding & Temporal (bidirectional time-delta attention) \\
    \quad Temporal encoding layers & 0-3 \\
    \quad Numerics Fourier basis & 22 dyadic scales ($2^{-7}$-$2^{14}$) \\
    \quad LayerScale init & $1\times 10^{-4}$ \\
    \midrule
    \addlinespace[2pt]
    \multicolumn{2}{@{}l}{\textit{Optimization}} \\
    \addlinespace[1pt]
    \quad Optimizer & Muon (hidden matrices) + AdamW (embeddings, heads, non-2D) \\
    \quad Peak learning rate & $6\times 10^{-4}$ \\
    \quad Learning rate schedule & Cosine with linear warmup \\
    \quad Warmup epochs & 6 \\
    \quad Weight decay & 0.01 \\
    \quad Gradient clip norm & 10 \\
    \quad Dropout & 0.1 \\
    \quad Drop path & 0.1 \\
    \quad Attention dropout & 0.1 \\
    \midrule
    \addlinespace[2pt]
    \multicolumn{2}{@{}l}{\textit{Training Objective}} \\
    \addlinespace[1pt]
    \quad Prior-decoded reconstruction weight & 0.1 \\
    \quad Posterior-decoded reconstruction weight & 0.3 \\
    \midrule
    \addlinespace[2pt]
    \multicolumn{2}{@{}l}{\textit{Variational Latent}} \\
    \addlinespace[1pt]
    \quad KL weight $\beta_{\max}$ & 1 \\
    \quad KL warmup epochs (linear anneal) & 10 \\
    \quad Prior standard-deviation floor & 0.05 \\
    \quad Standard-deviation cap (smooth squash) & 2 \\
    \bottomrule
  \end{tabular}
\end{table}

\end{nolinenumbers}

\end{document}